\documentclass{article}

\usepackage{arxiv}

\usepackage[utf8]{inputenc}
\usepackage[T1]{fontenc}
\usepackage{hyperref}
\usepackage{url}
\usepackage{booktabs}
\usepackage{amsfonts}
\usepackage{nicefrac}
\usepackage[expansion=false]{microtype}
\usepackage{graphicx}
\usepackage{float}
\usepackage[section]{placeins}
\usepackage{amsmath,amssymb,amsthm}
\usepackage{multirow}
\usepackage{array}
\usepackage[dvipsnames]{xcolor}
\usepackage{caption}
\usepackage{enumitem}
\usepackage{subcaption}
\usepackage{cleveref}
\usepackage{setspace}
\usepackage{natbib}
\usepackage{doi}
\usepackage{adjustbox}
\usepackage{pifont}
\usepackage{tabularx}
\usepackage{mathrsfs}
\usepackage{bm}
\usepackage{colortbl}
\usepackage{textcomp}

\usepackage{algorithm}
\usepackage{algpseudocode}

\usepackage{tikz}
\usetikzlibrary{shapes.geometric, positioning, calc, fit}
\tikzset{myarrow/.style={-stealth, thick}}

\usepackage{siunitx}
\title{MARTS: A Modular Agricultural Robot for Targeted Spraying of Diseased Tomato Plants in Greenhouses}

\author{
  Rajmeet Singh Bhourji$^{*}$ \\
  Mohamed bin Zayed University of Artificial Intelligence\\
  Abu Dhabi, UAE \\
  Khalifa University Center for Autonomous Robotic Systems (KUCARS)\\
  Khalifa University, Abu Dhabi, UAE \\
  \texttt{rajmeet.bhourji@mbzuai.ac.ae}
  \And
  Manveen Kaur \\
  Department of Mechanical Engineering\\
  University of Windsor\\
  ON, Canada \\
  \texttt{Kaur79@uwindsor.ca}
  \And
  Yusuf Adamu \\
  Khalifa University Center for Autonomous Robotic Systems (KUCARS)\\
  Khalifa University, Abu Dhabi, UAE \\
  \texttt{yusuf.adamu@ku.ac.ae}
  \And
  Lochan Kshetrimayum \\
  Kobe University\\
  Japan \\
  \texttt{lochan@ku.ac.ae}
  \And
  Irfan Hussain \\
  Khalifa University Center for Autonomous Robotic Systems (KUCARS)\\
  Khalifa University, Abu Dhabi, UAE \\
  \texttt{irfan.hussain@ku.ac.ae}
  \And
  \vspace{-1em}
  \normalsize\textit{$^{*}$ Corresponding author.}
}

\hypersetup{
  pdftitle={MARTS: A Modular Agricultural Robot for Targeted Spraying of Diseased Tomato Plants in Greenhouses},
  pdfsubject={Agricultural Robotics, Precision Spraying, Vision-Language Models, Greenhouse Automation},
  pdfauthor={Rajmeet Singh Bhourji, Manveen Kaur, Yusuf Adamu, Lochan Kshetrimayum, Irfan Hussain},
  pdfkeywords={Agricultural robotics, Selective spraying, Greenhouse automation, Tomato disease management, Modular end-effector, Vision-language model, UR5 manipulator, Husky A200, ROS},
  colorlinks=true,
  citecolor=blue,
  linkcolor=blue,
  urlcolor=blue
}

\begin{document}
\maketitle

% ── Abstract ──────────────────────────────────────────────────────────────────
\begin{abstract}
Broadcast spraying remains the dominant method of chemical disease control in greenhouse tomato production, yet it applies pesticide indiscriminately to healthy and diseased tissue, exposes human applicators to hazardous aerosols, and delivers variable treatment quality that depends on operator skill and fatigue. This paper presents \textbf{MARTS} (\textbf{M}odular \textbf{A}gricultural \textbf{R}obot for \textbf{T}argeted \textbf{S}praying), a low-cost robotic platform that closes the loop from plant level disease detection to precise, leaf localised spray delivery inside commercial greenhouses. The platform integrates a Clearpath Husky~A200 differential-drive mobile base, a Universal Robots UR5 six-axis collaborative manipulator, and a novel modular low cost spraying end-effector built around a single 12~V~DC micro-diaphragm pump, and an Intel RealSense D435 RGB-D camera mounted on the end-effector for wrist level perception. Disease detection is performed by the previously published Tomato Leaf Disease Vision Language Model (TLDVLM), a three-stage pipeline comprising Grounding DINO for open-vocabulary leaf localisation, SAM-2 for pixel-precise lesion segmentation, and BLIP-2 with Low-Rank Adaptation fine-tuned across 10 tomato leaf disease classes at 97.27\% classification accuracy; the present contribution reuses this model as a drop-in perception front-end and focuses on the downstream actuation, control, and system integration. A ROS-based detection to spray pipeline lifts pixel space bounding-box detections into 3D manipulator targets through the RGB-D geometry, generates collision aware nozzle approach poses at a calibrated stand-off of 120~mm using fitted leaf surface normals, and triggers timed pump activations whose duration is scaled to the detected lesion area via the model $\tau = \tau_0 + k\,A$ ($\tau_0 = 0.25$~s, $k = 1.0 \times 10^{-4}$~s/mm$^2$). Bench calibration on water-sensitive paper at three stand off distances (80, 120, 160~mm) established a linear pulse-duration volume relationship ($R^2 = 0.994$) and identified 120~mm as the optimal coverage overspray operating point. Prior to hardware deployment, trajectory planning was validated in a MuJoCo simulation across three mobile-base positioning strategies (frontal, diagonal, and adaptive repositioning); the Case~III adaptive strategy achieved 100\% workspace coverage of 18 predefined waypoints with a maximum orientation error of only 1.1$^{\circ}$, and was selected for field deployment. The physical platform was then deployed at the SILAL commercial greenhouse facility (Al~Ain, Abu Dhabi, UAE) on three rows of mature cherry tomato plants under natural disease conditions. Across 312 detected diseased leaves treated in 12 autonomous trials, MARTS achieved a targeting success rate of \textbf{92.4\%}, a mean lateral spray-centre error of \textbf{18.7~mm} (SD~7.4~mm; 95th-percentile 32.1~mm) relative to the lesion centroid, an adjacent-leaf overspray rate of only \textbf{6.1\%}, and a \textbf{71\%} reduction in pesticide volume relative to a calibrated broadcast baseline at comparable diseased-tissue coverage. The average per leaf cycle time from detection to spray completion was \textbf{4.8~s}, dominated by the velocity limited manipulator approach (3.1~s). The total bill of materials for the spraying subsystem is \textbf{USD~412}, which is 6--10$\times$ lower than comparable systems reported in the literature, demonstrating that competitive leaf level targeting accuracy can be delivered on commodity hardware. This work establishes a practical framework that bridges vision language perception with low-cost, modular spraying hardware for autonomous, damage free disease management in commercial greenhouse tomato production.
\end{abstract}


% ═════════════════════════════════════════════════════════════════════════════
\section{Introduction}
\label{sec:introduction}
% ═════════════════════════════════════════════════════════════════════════════

Plant diseases pose a serious risk to global food security, with damaging consequences for staple crops such as wheat and tomato that underpin agricultural economies worldwide. In Canada, greenhouse production alone yields around 294{,}000 tonnes of tomatoes each year across more than 6{,}200~ha of cultivated area, generating revenues that exceed 600~million~CAD \citep{statcan2023greenhouse}. The United Arab Emirates (UAE), meanwhile, has rapidly transformed its tomato sector through the deployment of controlled-environment agriculture (CEA) and advanced greenhouse technologies an approach driven by the need to overcome the country's arid conditions, extreme summer temperatures, and acute scarcity of freshwater resources \citep{moccae2021agstrategy}. Within these protected environments, yield, fruit quality, and shelf life are maintained primarily through routine chemical disease management, with foliar pathogens such as early blight (\textit{Alternaria solani}), late blight (\textit{Phytophthora infestans}), powdery mildew (\textit{Leveillula taurica}), and bacterial spot \citep{ren2020review, oberti2016selective} controlled by weekly or sub-weekly fungicide and bactericide applications during the symptomatic phase of the cropping cycle.

\begin{figure}[!ht]
    \centering
    \includegraphics[width=0.5\linewidth]{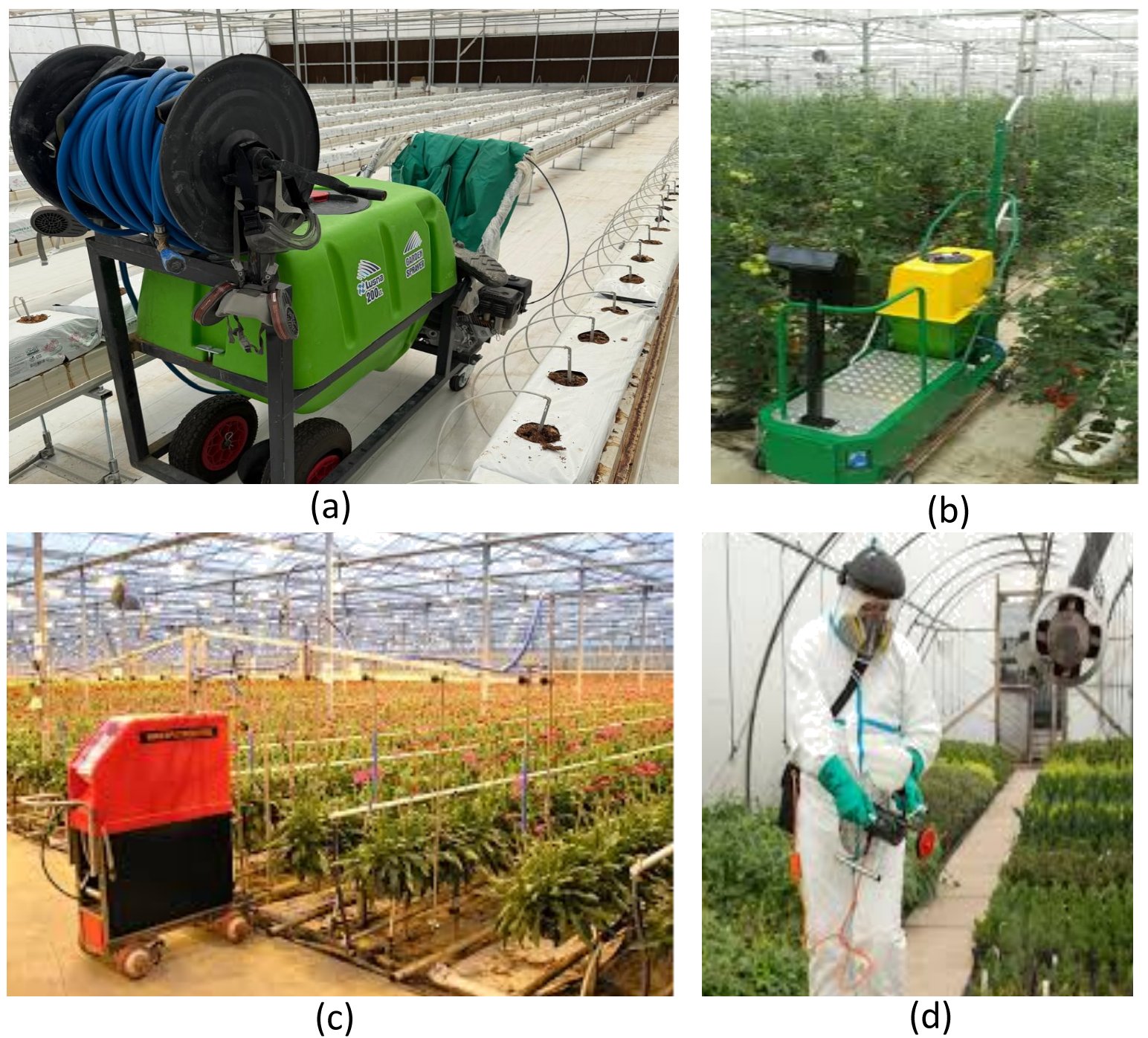}
    \caption{Representative conventional bulk-spraying practices in greenhouse production. (a) Hose reel trolley sprayer with 200~L tank for manual broadcast application along the row. (b) Self-propelled compact sprayer for ornamental and vegetable greenhouses. (c) Operator pushed cart sprayer with vertical nozzle boom for uniform canopy treatment. (d) Pesticide applicator in full PPE chemical resistant coverall, respirator, face shield, and gloves.}
    \label{fig:fig1}
\end{figure}

Despite the technological sophistication of modern greenhouses, the dominant method of pesticide application in commercial tomato production remains \emph{bulk}, or \emph{broadcast}, spraying, chemicals are applied uniformly across the entire canopy through manually carried knapsack sprayers, motorised mist blowers, or boom-mounted spraying rigs \citep{lochan2024advancements, oberti2016selective} as depicted in Fig.~\ref{fig:fig1}. This practice is operationally simple but agronomically blunt: chemicals are deposited indiscriminately on healthy and diseased tissue alike, a large fraction of the spray volume drifts to the soil, the structural walls, and the trellising hardware, and the applied dose is sized for the worst-case disease pressure rather than the actual symptomatic load \citep{kootstra2021selective, gonzalez2016autonomous}. Several practical and human-factor limitations of bulk spraying have been documented in the recent literature, and they constitute the central motivation of the present work.

Bulk spraying inside an enclosed greenhouse volume exposes the human applicator to airborne pesticide droplets, respirable aerosols, and re-entrained chemical residues at concentrations substantially higher than those encountered in open-field application \citep{damalas2011pesticide, kim2017pesticide, yarpuz2018ppe}. Documented effects include acute symptoms (headache, nausea, dermal and ocular irritation) as well as chronic outcomes associated with long-term, low-dose exposure to organophosphate, carbamate, and triazole compounds \citep{abdollahi2004pesticides, rauh2012brain, burns2013pesticide}. In response, regulatory frameworks worldwide mandate that bulk-spraying operators wear full personal protective equipment (PPE) consisting of chemical resistant coveralls, nitrile gloves, rubber boots, a face shield, and a respirator with appropriate cartridges. Even when correctly worn, PPE imposes substantial physiological load on the operator, particularly in high-temperature, high-humidity greenhouse environments typical of CEA installations in the Gulf region, where ambient temperatures inside the protected volume can exceed 35\,$^{\circ}$C during much of the cropping season \citep{wachenheim2021uav, baltazar2021smarter}. PPE compliance among small and medium growers also remains inconsistent in practice, further amplifying occupational exposure risk \citep{elhalwagy2010knapsack}.

Effective bulk spraying is not merely a matter of opening a valve; it requires the operator to maintain a calibrated walking pace, a consistent nozzle stand-off from the canopy, a uniform overlap between adjacent passes, and a correctly adjusted nozzle pressure all while wearing restrictive PPE in a hot, humid environment \citep{lochan2024advancements, oliveira2021advances}. Under-trained operators routinely produce under-treated zones (which propagate the disease) and over-treated zones (which waste chemicals and exacerbate residue on the harvest) \citep{wachenheim2021uav}. The labour pool capable of performing this task to a consistent standard is shrinking in both developed and developing agricultural economies, as greenhouse operators increasingly compete for skilled labour against construction, logistics, and service sectors that offer better working conditions \citep{ishii2021tomato, rapado2023development, bogue2024robots}. The combined effect, rising wages, shrinking labour availability, and growing reluctance to perform chemical-handling work has made manual bulk spraying one of the most cost-pressured operational tasks in commercial tomato production. Manual bulk spraying is inherently variable from operator to operator, from pass to pass, and even within a single pass as the applicator fatigues. The resulting heterogeneity in spray deposit on the diseased tissue is a recognised cause of repeated regrowth of the infection \citep{lochan2024advancements, gonzalez2016autonomous}.

The limitations above motivate the shift from bulk spraying to \emph{targeted}, or \emph{selective}, spraying, in which the chemical is delivered only where and when it is needed, and at a dose sized to the local symptomatic load \citep{west2003potential, oberti2016selective, lochan2024advancements}. Targeted spraying is operationally attractive on three counts simultaneously: it removes the human applicator from the chemical handling loop (eliminating PPE compliance as a daily occupational issue), it reduces the total chemical load per cropping cycle (with direct cost and environmental benefits), and it provides a repeatable, operator independent treatment quality that bulk spraying cannot deliver.

Two broad classes of agricultural robotic platforms have emerged in support of this paradigm. \emph{Aerial} platforms, typically multirotor unmanned aerial vehicles (UAVs), have been extensively investigated for open-field row-crop applications \citep{huang2009development, faical2017adaptive, xue2016develop, lochan2024advancements}. UAVs are effective for canopy-level coverage at field scale, but they are poorly suited to greenhouse tomato cultivation because the enclosed volume, the high-wire trellising, and the dense inter-row foliage do not provide the rotor clearance and the flight envelope that UAV operation requires, and because the canopy underside, where many tomato pathogens express symptomatically is unreachable from above \citep{lochan2024advancements, oberti2016selective}. \emph{Terrestrial} ground-based platforms, in contrast, can navigate the narrow aisles of commercial greenhouses and can position a manipulator mounted nozzle at the leaf scale, including on the underside of the canopy. Several research prototypes have demonstrated the technical feasibility of this approach \citep{oberti2016selective, baltazar2021smarter, cantelli2019small, rincon2020spray, sammons2005autonomous, gonzalez2016autonomous, hejazipoor2021intelligent}, and a smaller number of commercial platforms (e.g.\ ecoRobotix AVO, Sitia Trektor) have begun to enter the market \citep{bykov2023world}.

Despite the visible activity in this space, three gaps in the current literature continue to limit the practical adoption of robotic targeted spraying in commercial greenhouse tomato production. First, most ground manipulator selective sprayers reported in the literature are built on research-grade platforms whose total cost particularly the spraying subsystem places them beyond the reach of commercial growers operating on tight per hectare budgets \citep{lochan2024advancements, baltazar2021smarter}. The robotic spraying literature is rich in demonstrations on expensive, purpose built hardware, but comparatively sparse in designs that explicitly target a low marginal cost above the underlying mobile manipulator. Second, most published selective spraying systems report \emph{detection} accuracy but not \emph{spraying} accuracy the spatial distance between the centre of the deposited spray pattern and the centroid of the diseased lesion, and the fraction of detected lesions that receive a physical spray pulse. Without this quantitative actuation characterisation, the chain from a detection event to a physical, correctly sized, correctly aimed spray remains effectively unverified \citep{oberti2016selective, partel2021smart}. Third, existing spraying end effectors are typically purpose built for a specific nozzle, chemical, and crop, with the spraying subsystem mechanically and electrically intertwined with the manipulator's tool flange; there is no widely reported modular design that allows the grower to swap nozzles and chemicals between tomato, cucumber, and pepper without rebuilding the manipulator's mechanical or control stack.

\subsection{Research gaps}

Three gaps in the current literature motivate this work. First, robotic spraying end-effectors are typically built on expensive, purpose built hardware and are rarely designed as independent, low-cost, modular subsystems \citep{lochan2024advancements, baltazar2021smarter}. Second, vision language models (VLMs) for tomato disease detection have advanced rapidly, but their integration with a downstream selective spraying actuator has not been demonstrated end-to-end on a real greenhouse platform \citep{oberti2016selective, partel2021smart}. Third, most published systems are validated either in simulation or in the field, but rarely in both. This work addresses all three gaps through (i)~a low-cost modular spraying end-effector (BOM~$<$~USD~420), (ii)~a VLM-guided selective-spraying pipeline that reuses a previously published tomato leaf disease detector \citep{kaur2025visual}, and (iii)~end-to-end validation in MuJOCO and at the SILAL commercial greenhouse, Abu Dhabi, UAE.

\subsection{Contributions}

The principal contributions of this work are therefore fivefold:
\begin{enumerate}[leftmargin=*, itemsep=2pt]
    \item A low-cost, modular spraying end-effector (BOM~$<$~USD~420) with a 3D-printed tool-flange adapter and a quick-change nozzle interface enabling tip replacement.

    \item A quantitative spray actuation model ($\tau = \tau_0 + k\,A$) calibrated on water-sensitive paper and validated at three stand-off distances, providing a principled mapping from detected lesion area to spray pulse duration.

    \item A complete ROS-based detection-to-spray pipeline that reuses a previously published VLM disease detector as a drop-in perception front-end, demonstrating the practical value of decoupling perception from actuation in agricultural robotics.

    \item A MuJoCo-based trajectory planning (three positioning strategies, 18 waypoints) with quantitative sim-to-real transfer confirmation in the greenhouse.

    \item End-to-end experimental validation under genuine commercial greenhouse conditions, with quantitative reporting of targeting success, spatial accuracy, overspray, volume reduction, cycle time, and failure mode analysis across 312 autonomous spraying events.
\end{enumerate}

% ═════════════════════════════════════════════════════════════════════════════
\section{Materials and Methods}
\label{sec:materials}
% ═════════════════════════════════════════════════════════════════════════════

\subsection{System Overview}

\begin{figure}[!ht]
    \centering
    \includegraphics[width=0.5\linewidth]{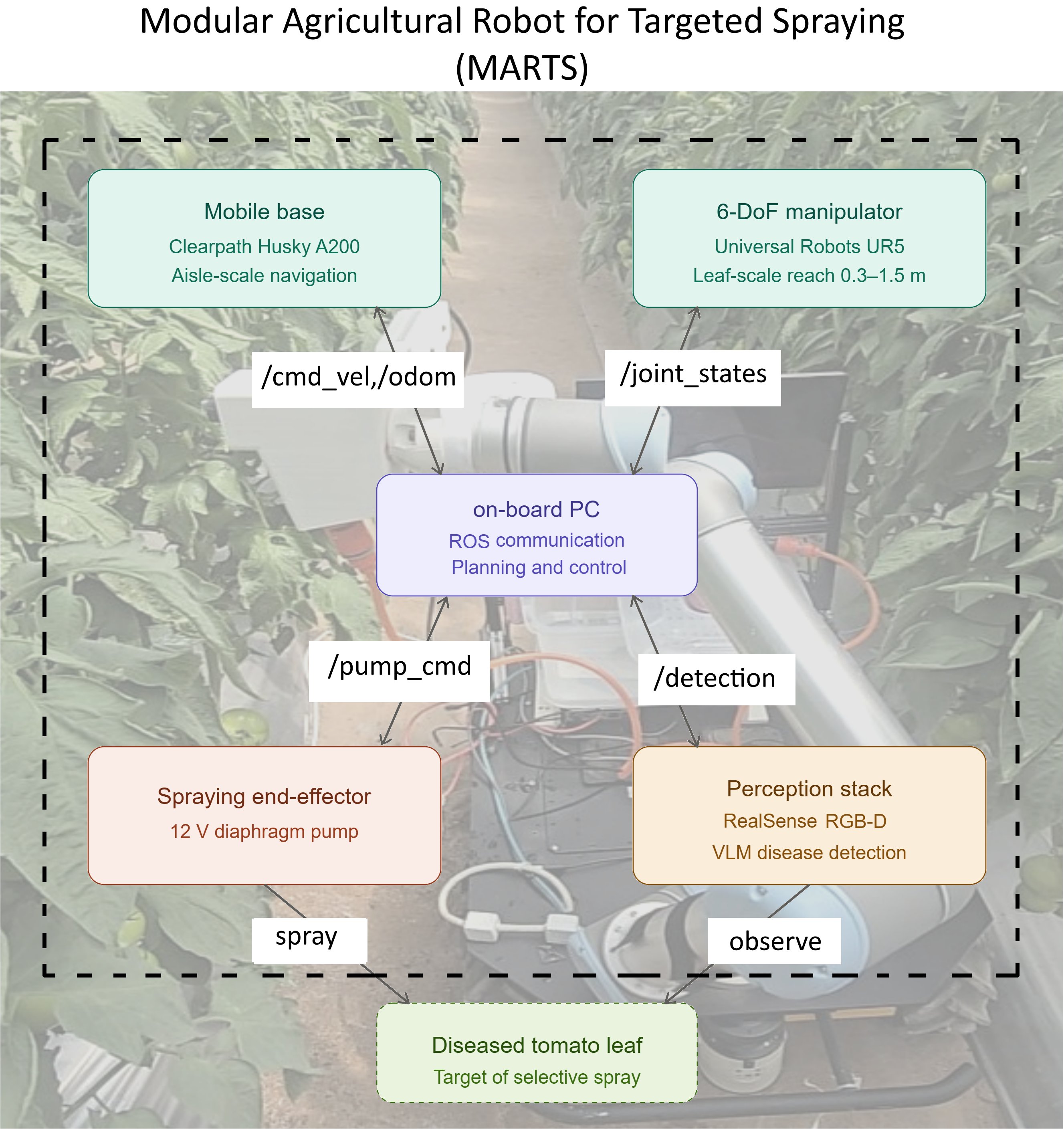}
    \caption{System architecture of the MARTS platform. The four subsystems: Clearpath Husky~A200 mobile base, Universal Robots UR5 manipulator, modular spraying end-effector, and on-board perception stack with the Intel RealSense D455 RGB-D camera communicate through a ROS graph running on a single on-board PC. The dashed green target indicates the diseased tomato leaf: the perception stack observes it through the VLM disease detector, and the end-effector delivers the leaf-localised spray pulse, closing the detection-to-spray loop.}
    \label{fig:fig2}
\end{figure}

MARTS comprises four subsystems: a mobile base, a 6-DoF manipulator, a modular spraying end-effector, and an on-board perception and compute stack. The mobile base provides aisle-scale navigation between plant rows. The manipulator provides plant scale reach to individual leaves at heights between approximately 0.3~m and 1.5~mm, covering the symptomatic range of mature greenhouse tomato. The spraying end-effector converts a commanded spray duration and approach pose into a metered chemical delivery. The perception stack provides disease detections. All subsystems communicate over ROS architecture on a single on-board husky PC. Fig.~\ref{fig:fig2} presents the complete MARTS system architecture, showing the hardware layout and the data flow between subsystems.

The mobile base is a Clearpath Husky A200, a four-wheeled differential drive platform providing a maximum payload capacity of 75~kg, a top speed of 1.0~m/s, and integrated on-board computing. The Husky is equipped with a SICK LMS151 2D LiDAR (270$^{\circ}$ field of view, 50~m range) for aisle-following and obstacle detection, and wheel odometry fused with an IMU through the standard \texttt{robot\_localization} package. In the greenhouse trials reported here, the Husky operated in a teleoperated waypoint mode between rows; the autonomous navigation stack is operational but is not the subject of this paper. The Husky's payload capacity comfortably accommodates the manipulator, the chemical reservoir, spraying pump, and the compute stack. The UR5 manipulator is installed on husky platform and communicate with On-board PC via IP communication. The low cost spraying mechanism + nozzle (end effector) is also mounted on the Husky platform. The detail description of end effector design and spraying mechanism is mentioned in next section.

\subsection{Modular Spraying End-Effector}

\begin{figure}[!ht]
    \centering
    \includegraphics[width=0.75\linewidth]{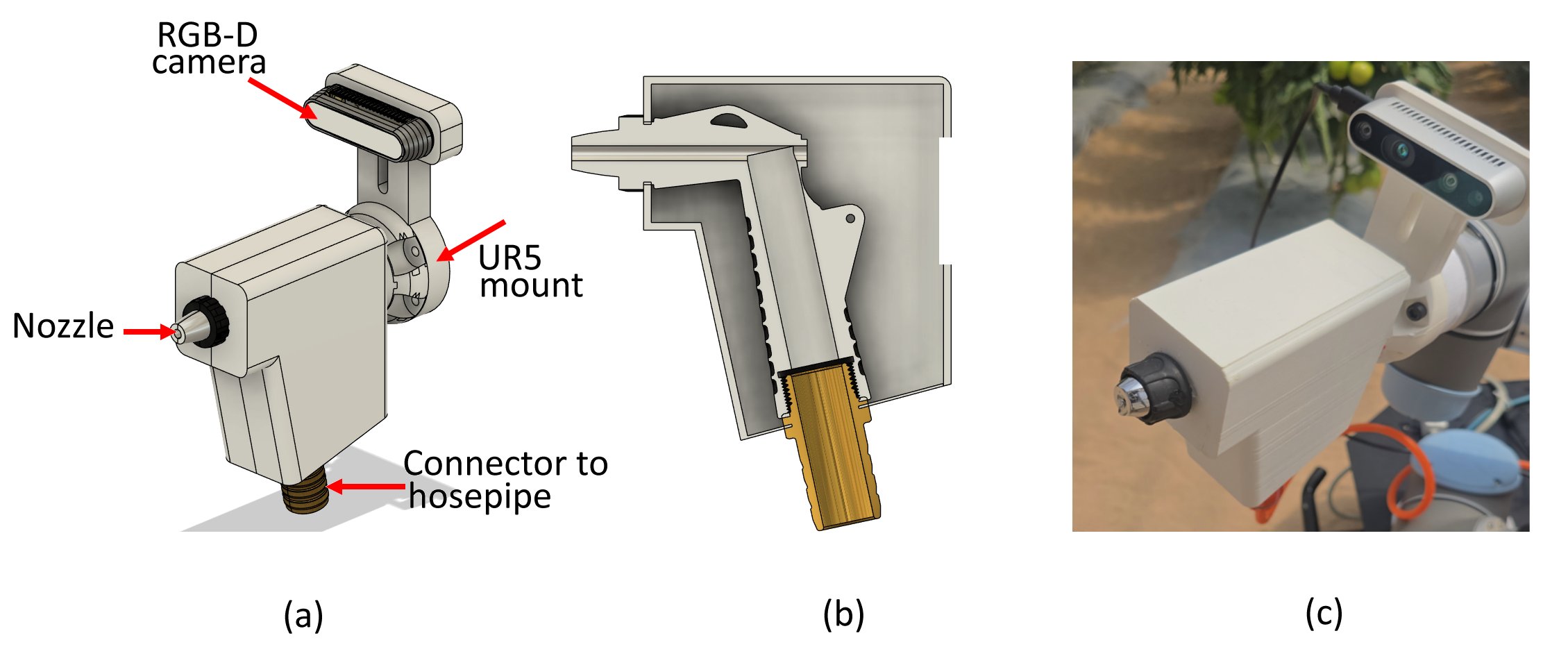}
    \caption{Selective spraying end-effector designed for integration with the UR5 manipulator: (a) isometric CAD rendering showing the key components, including the Intel RealSense RGB-D camera mounted on top for target localization, the spraying nozzle at the front, the UR5 tool-flange mount, and the hosepipe connector at the base; (b) sectional cutaway view illustrating the internal routing of the brass threaded fitting that channels the spray fluid from the hosepipe connector to the nozzle outlet; (c) 3D-printed prototype mounted on the UR5 end-effector during field deployment in the greenhouse.}
    \label{fig:fig3}
\end{figure}

The end-effector is the central hardware contribution of this paper and is designed under three constraints: low cost, modular reconfiguration without rebuilding the manipulator's tool flange, and operational safety inside an enclosed greenhouse volume. Fig.~\ref{fig:fig3} illustrates the complete end-effector assembly, and the integration with the UR5 tool flange.

The precision spraying end effector was a device designed to generate a pesticide liquid spray to deliver spraying patterns on target patches. Table~\ref{tab:table1} mentioned the specifications of the spraying mechanism. A critical advantage of this design is its economic accessibility: the complete spraying end-effector assembly, including the pump and water tank, costs less than USD 500, significantly lowering the barrier to adoption for small scale and resource constrained farming operations.

\begin{table}[!ht]
\centering
\caption{Modular spraying end-effector specifications.}
\label{tab:table1}
\small
\renewcommand{\arraystretch}{1.2}
\begin{tabular}{lc}
\toprule
\textbf{Parameter} & \textbf{Value} \\
\midrule
Pump type                 & 12~V DC diaphragm \\
Free-flow rate            & 4.0~L/min \\
Max pressure              & 0.55~MPa \\
Reservoir capacity        & 10.0~L (food-grade) \\
Tubing inner diameter     & 4.0~mm (polyurethane) \\
Nozzle type               & Hollow-cone (default) \\
Nozzle spray angle        & 60$^{\circ}$ \\
Nominal flow @ 0.3~MPa    & 0.32~L/min \\
Interchangeable tips      & 5 (hollow-cone, flat-fan, etc.) \\
Pump control              & arduino, ROS topic \\
\bottomrule
\end{tabular}
\end{table}

\subsubsection{System Architecture and Spray Actuation}

\begin{figure}[!ht]
    \centering
    \includegraphics[width=\linewidth]{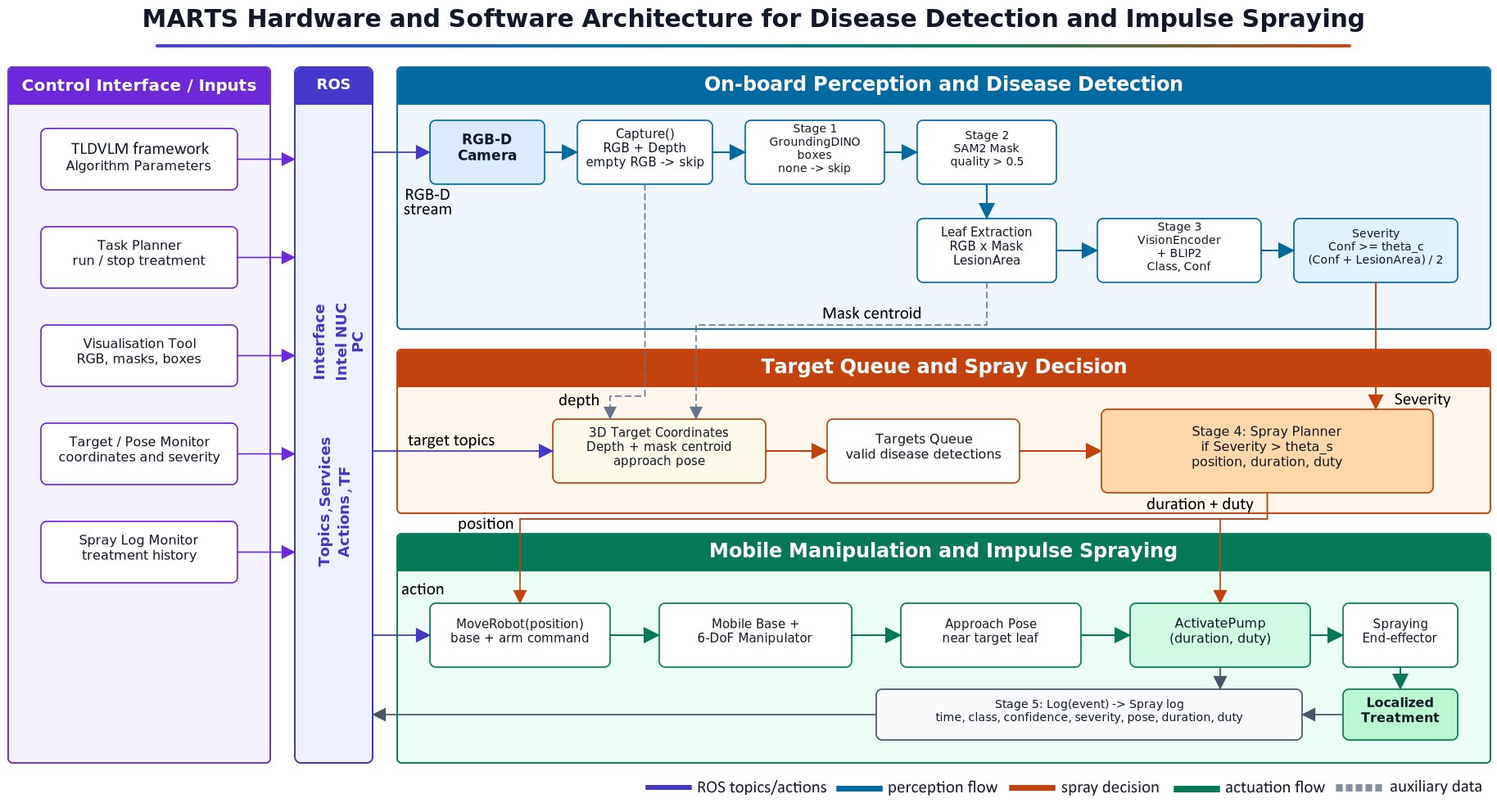}
    \caption{MARS integrated hardware-software architecture for autonomous disease detection and selective impulse spraying. The system orchestrates perception (GroundingDINO, SAM-2, BLIP-2~LoRA), spray decision logic, and mobile manipulation via ROS~2.}
    \label{fig:fig4}
\end{figure}

The spraying system integrates ROS~2 based supervisory control with multi-stage perception and low level hardware actuation. Figure~\ref{fig:fig4} illustrates the complete hardware-software architecture, which is organized into five functional stages: on-board perception, target queue management, spray decision logic, mobile manipulation, and comprehensive treatment logging.

\textbf{Perception and Disease Detection Pipeline:} Live RGB-D streams from an Intel RealSense~D435 mounted on the end-effector flow through a multi-stage deep-learning architecture: (Stage~1) GroundingDINO performs open vocabulary leaf region detection; (Stage~2) SAM-2 applies pixel precise lesion segmentation with quality filtering (mask quality threshold $\theta_{\text{mask}} > 0.5$); (Stage~3) a VisionEncoder with BLIP-2 LoRA fine-tuned on 10 tomato leaf disease classes outputs disease classification and confidence scores; (Stage~4) disease severity is computed as $\text{Severity} = (\text{Confidence} + \text{LesionArea})/2$, thresholded at confidence $\theta_c = 0.70$. This integrated approach facilitates closed-loop control where visual perception of disease severity and spatial location directly informs subsequent actuation decisions.

\textbf{Target Queue and Spray Decision Logic:} Perception outputs are transformed into 3D actionable targets via depth-to-world coordinate transformation using the mask centroid and depth values. The Target Queue applies severity based filtering ($\theta_s > 0.75$) to prioritize affected leaf patches and avoid redundant spraying of healthy foliage. The Spray Decision module converts severity metrics into temporal spray parameters: low severity (500~ms duration, 0.40 duty cycle), medium severity (750~ms, 0.60 duty cycle), and high severity (1000~ms, 0.80 duty cycle), ensuring pesticide dosage is proportional to disease intensity.

\begin{figure}[!ht]
    \centering
    \includegraphics[width=0.5\linewidth]{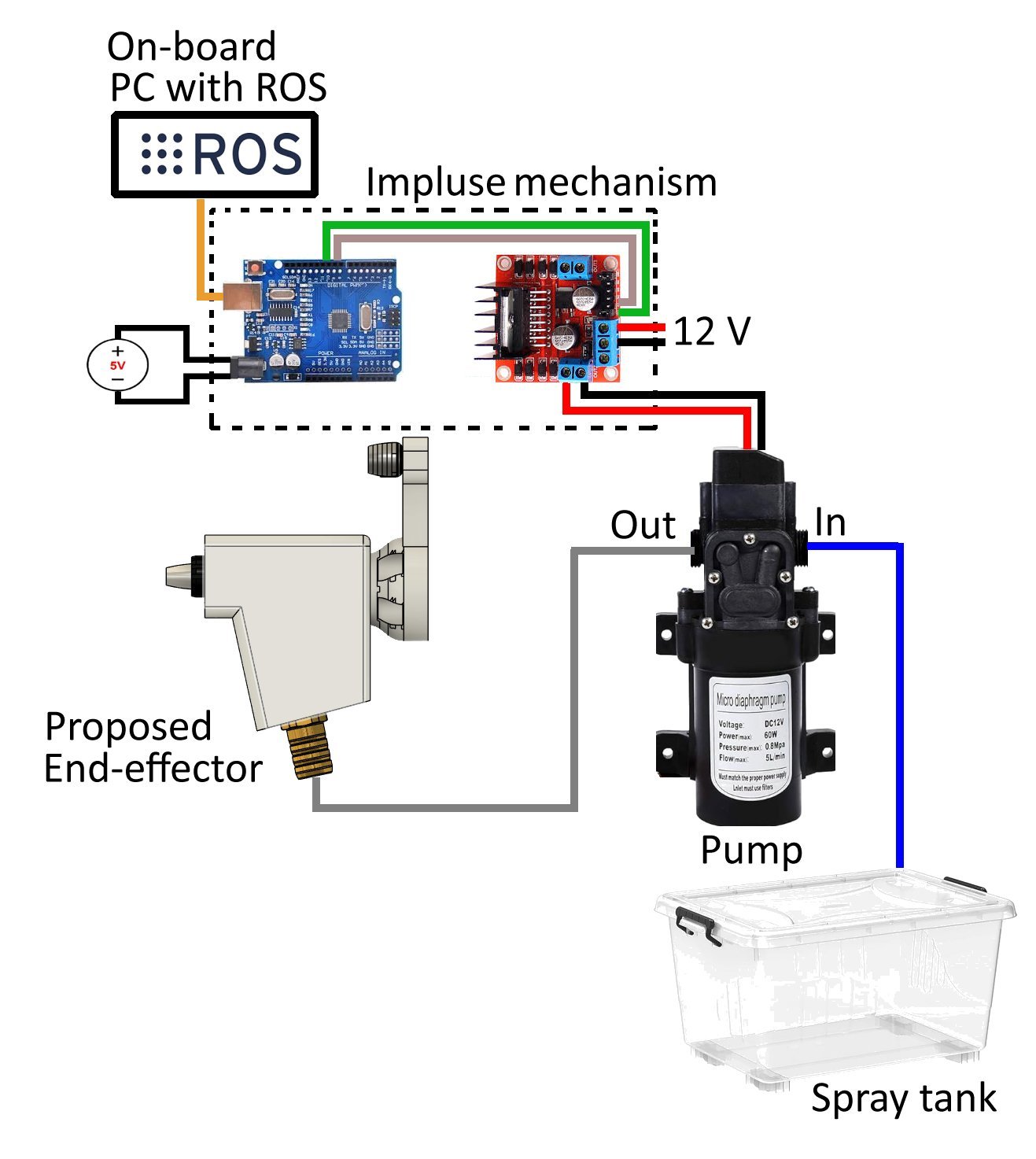}
    \caption{System architecture of the precision spraying platform: the on-board ROS-based PC communicates via impulse mechanism with an Arduino-controlled pump driver (5 V input) to command the micro-diaphragm pump (12 V supply). The pump draws liquid from the spray tank and delivers pressurized fluid to the proposed end-effector, which atomizes the pesticide formulation for selective spot-spraying application.}
    \label{fig:fig5}
\end{figure}

\textbf{Hardware Actuation and Coordination:} An Arduino-controlled impulse mechanism commands a 12~V micro-diaphragm pump (5~W, 5~L\textperiodcentered min$^{-1}$ rated capacity) to draw pesticide formulation from a 20~L tank and deliver it under pressure to the end-effector nozzle. This impulse driven approach enables precise, energy-efficient spray pulses synchronized with real-time target detection from the RGB-D camera, facilitating selective application on affected plant regions. The mobile base and 6-DoF manipulator coordinate synchronized motion, executing approach trajectories to position the end-effector near target leaves while maintaining optimal viewing and spraying angles. Figure~\ref{fig:fig5} illustrates the wire schematic diagram of the selective spraying mechanism. Integration of the RGB-D sensor on the end-effector enables in-situ disease diagnosis, allowing the system to identify symptomatic leaf patches during autonomous scanning operations and execute selective spraying without human intervention.

\textbf{System Performance and Logging:} ROS~2 middleware orchestrates real-time communication and synchronization across all subsystems, with end-to-end perception-decision-actuation latency of 380--500~ms/frame at 10~fps sensor refresh rate. Comprehensive treatment metadata time stamps, classification confidence, lesion area, end-effector pose, spray duration, and duty cycle are logged for each spray event.

\subsection{Tomato Leaf Disease Sensing System}

\subsubsection{Background and Previous Work}

Earlier research by the authors developed a comprehensive Tomato Leaf Disease Visual Language Model (TLDVLM) specifically designed for portable greenhouse monitoring and disease detection applications \citep{kaur2025visual}. This prior work established the foundation for integrating advanced vision-based disease classification into robotic agricultural systems. The TLDVLM successfully demonstrated the feasibility of deploying state-of-the-art vision-language models on resource constrained hardware, achieving 97.27\% classification accuracy across 10 distinct tomato leaf disease classes while maintaining computational efficiency suitable for edge deployment.

\subsubsection{Proposed Disease Detection Integration}

Building upon the established TLDVLM framework, the current work integrates real-time tomato leaf disease detection as a sensor feedback component within the autonomous spraying platform as depicted in Algorithm~\ref{alg:algo1}. The Intel RealSense D435 RGB-D camera mounted on the end-effector serves a dual functionality: it enables precise disease identification and spatial localization of symptomatic leaf patches, thereby automating the decision making process for selective spray actuation. The disease detection pipeline operates through a three stage architecture comprising object detection (GroundingDINO for leaf localization), pixel-level segmentation (SAM-2 for precise lesion delineation), and multimodal disease classification (BLIP-2 with Low-Rank Adaptation for parameter efficient fine tuning). This integrated approach facilitates closed-loop control where visual perception of disease severity and spatial location directly triggers the impulse driven spray mechanism, ensuring pesticide delivery is confined exclusively to affected regions and minimizing chemical application on healthy foliage. The system thereby transforms the robotic spraying platform from a manually triggered device to a fully autonomous, perception driven agricultural system capable of real-time decision-making in greenhouse environments.

Each detection $b_i$ is lifted into 3D using the pinhole camera model of the D435. Let $(u_{c,i}, v_{c,i})$ be the pixel centre of bounding box $b_i$, $z_i$ the median depth over the inner 50\% of the box, and $(f_x, f_y)$ and $(c_x, c_y)$ the calibrated focal lengths and principal point. The leaf target in the camera frame is
\begin{equation}
\mathbf{p}_{c,i} =
\begin{bmatrix}
\dfrac{(u_{c,i} - c_x)\, z_i}{f_x} &
\dfrac{(v_{c,i} - c_y)\, z_i}{f_y} &
z_i
\end{bmatrix}^{\top},
\label{eq:lift}
\end{equation}
which is then transformed to the manipulator base frame via the hand--eye calibration $\mathbf{T}_b^c$.

\begin{algorithm}[!ht]
\caption{Detection-to-Spray Controller for Autonomous
Selective Spraying of Diseased Tomato Plants}
\label{alg:algo1}
\small
\begin{algorithmic}[1]
\Require RGB-D image stream $\{I_t, D_t\}$, VLM model $\mathcal{M}$,
stand-off distance $d_s = 120$~mm, base spray duration
$\tau_0 = 0.25$~s, lesion-area scaling coefficient $k$,
positional tolerance $\epsilon = 5$~mm
\Ensure Spraying outcome (\textsc{Success} / \textsc{Failure})

\vspace{2pt}
\State \textbf{// Stage 1: VLM disease detection}
\State $\{b_i, c_i\}_{i=1}^{N} \gets \mathcal{M}(I_t)$
\Comment{Bounding boxes with confidence scores}
\If{$N = 0$}
    \State \Return \textsc{Failure} (no diseased leaves detected)
\EndIf
\State Sort detections by confidence in descending order

\vspace{2pt}
\State \textbf{// Stage 2: Lift detections to 3D}
\ForAll{detected leaves $b_i$}
    \State $z_i \gets \text{median}(D_t[b_i^{\text{inner}}])$
    \Comment{Median depth in inner 50\% of box}
    \State $\mathbf{p}_{c,i} \gets \text{Eq.}~\ref{eq:lift}$
    \State $\mathbf{p}_{b,i} \gets \mathbf{T}_b^c \mathbf{p}_{c,i}$
\EndFor

\vspace{2pt}
\State \textbf{// Stage 3: Generate approach poses}
\ForAll{$\mathbf{p}_{b,i}$}
    \State $\hat{\mathbf{n}}_i \gets \text{PlaneFit}(D_t[b_i^{\text{inner}}])$
    \State $\mathbf{p}_{a,i} \gets \mathbf{p}_{b,i} + d_s \cdot \hat{\mathbf{n}}_i$
    \State Set nozzle orientation anti-parallel to $\hat{\mathbf{n}}_i$
\EndFor

\vspace{2pt}
\State \textbf{// Stage 4: Motion planning and execution}
\ForAll{$\mathbf{p}_{a,i}$}
    \State $\mathbf{q}_i \gets \text{MoveIt\_Plan}(\mathbf{p}_{a,i})$
    \State \textsc{UR5\_Execute}($\mathbf{q}_i$)
    \State \textbf{wait until} $\|\mathbf{p}_{\text{actual}} - \mathbf{p}_{a,i}\| < \epsilon$

    \vspace{2pt}
    \State \textbf{// Stage 5: Compute and execute spray}
    \State $A_i \gets \text{Area}(b_i)$
    \State $\tau_i \gets \tau_0 + k \cdot A_i$
    \State \textsc{Pump\_On}()
    \Comment{Publish \texttt{True} to pump topic}
    \State \textbf{sleep}($\tau_i$)
    \State \textsc{Pump\_Off}()
    \Comment{Publish \texttt{False} to pump topic}
    \State \textsc{UR5\_Retract}()
\EndFor
\State \Return \textsc{Success} (all detected leaves treated)
\end{algorithmic}
\end{algorithm}

% ═════════════════════════════════════════════════════════════════════════════
\section{Plant Scanning Trajectory Planning}
\label{sec:trajectory}
% ═════════════════════════════════════════════════════════════════════════════

\begin{figure}[!ht]
    \centering
    \includegraphics[width=0.9\linewidth]{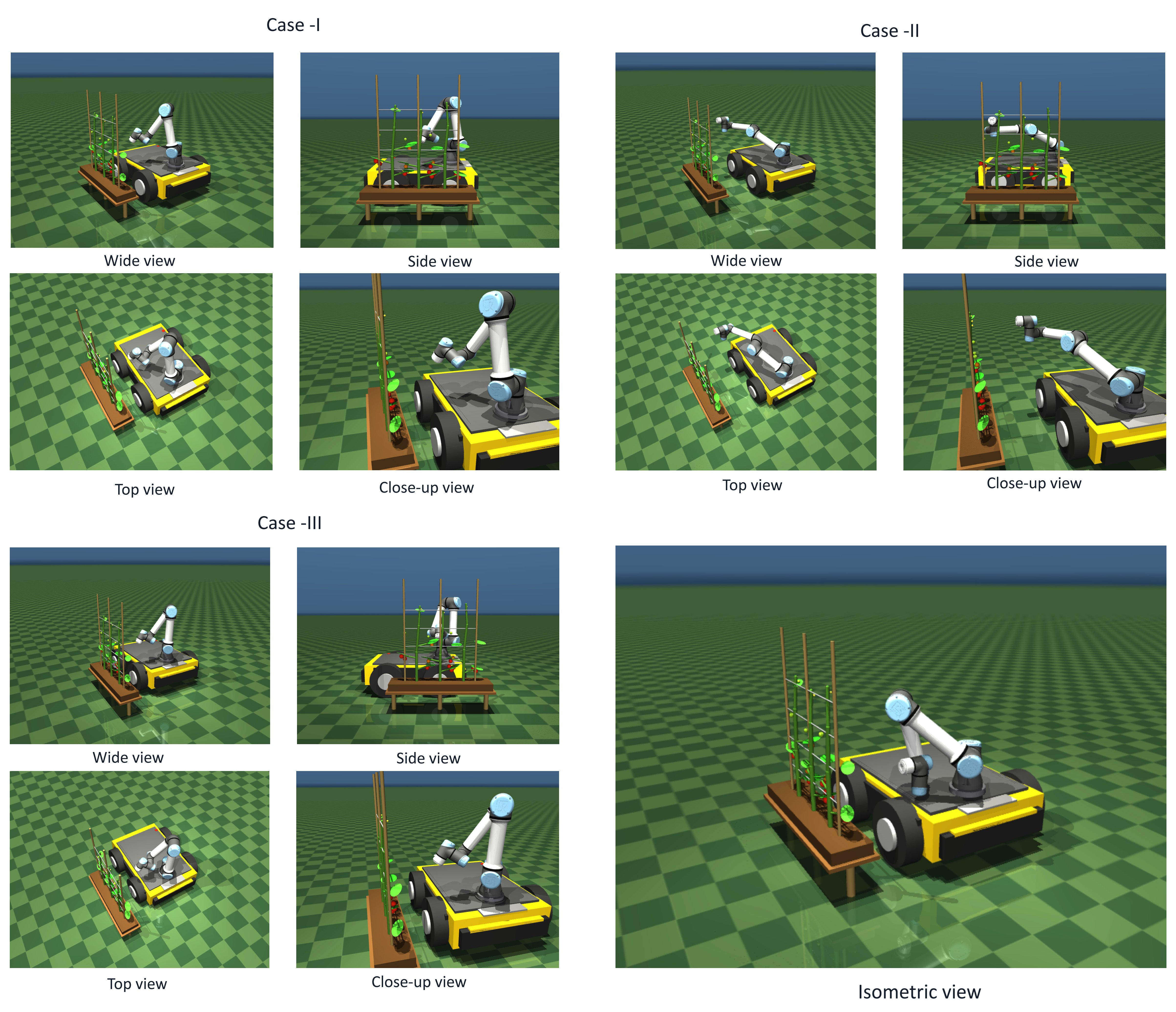}
    \caption{MuJoCo-based simulation framework for validating trajectory planning and mobile-base positioning strategies prior to hardware deployment.}
    \label{fig:fig7}
\end{figure}

Prior to real hardware implementation, comprehensive simulation studies were conducted to validate the autonomous plant scanning and spraying system in a controlled virtual simulation environment. This section presents the trajectory planning methodology, the MuJoCo based simulation framework to validate the trajectory tracking during different mobile base positioning strategies to determine optimal end-effector workspace accessibility and task performance as depicted in Fig.~\ref{fig:fig7}.

\subsection{Simulation Environment and Virtual tomato Plant Model}

\subsubsection{MuJoCo Physics Engine Configuration}

The trajectory planning and validation were performed using the MuJoCo 2.3 physics simulator, which provides high fidelity rigid body dynamics, accurate collision detection, and real-time computation necessary for mobile manipulation simulation. The simulation was configured with:

\begin{itemize}[leftmargin=*, itemsep=2pt]
    \item \textbf{Time step:} 0.001~s for accurate integration of joint and contact dynamics
    \item \textbf{Gravity:} 9.81~m/s$^2$ acting vertically downward
    \item \textbf{Collision detection:} MuJoCo's convex mesh algorithm for computationally efficient and physically accurate contact modeling
    \item \textbf{Contact dynamics:} Soft contact model with friction coefficient $\mu = 0.8$ for plant arm interactions and $\mu = 1.2$ for wheel ground traction
    \item \textbf{Solver:} Projected Newton-type solver with 4 iterations per time step
\end{itemize}

\subsubsection{Virtual Tomato Plant Environment}

A detailed virtual tomato plants were constructed to accurately represent the physical environment in which the MARTS system operates. The simulation environment includes:

\begin{enumerate}[leftmargin=*, itemsep=2pt]
    \item \textbf{Structural elements:} Greenhouse ground plane (concrete texture, 10~m $\times$ 5~m), raised planting beds constructed of wood (0.6~m height, 0.4~m width), support structures, and overhead irrigation lines modeled as rigid obstacles.

    \item \textbf{Plant models:} Artificial tomato plants arranged in regular grid patterns (0.3~m spacing between plants) on raised beds. Each plant model comprises:
    \begin{itemize}
        \item Flexible stem segment (modeled as a series of linked cylinders, 1.2--1.5~m height)
        \item Branch segments protruding at multiple levels
        \item Leaf geometry represented as flat rectangular meshes with contact properties
        \item Collision primitives assigned to all plant components to detect unwanted arm-foliage contact
    \end{itemize}

    \item \textbf{Mobile base:} Husky UGV model with UR5 6-DOF collaborative robot:
    \begin{itemize}
        \item Mass: 90~kg (including onboard computational equipment)
        \item Suspension: Simulated skid-steer kinematics with $\pm 0.15$~m/s linear velocity capability and $\pm 1.0$~rad/s angular velocity capability
        \item Manipulator arm link masses and inertias matching hardware specifications
        \item Joint position limits: $\theta_1 \in [-\pi, \pi]$, $\theta_2 \in [-\pi, \pi]$, $\theta_3 \in [-\pi, \pi]$, $\theta_4 \in [-\pi, \pi]$, $\theta_5 \in [-\pi, \pi]$, $\theta_6 \in [-\pi, \pi]$ rad
        \item Joint velocity limits: 180$^{\circ}$/s for all joints
        \item Torque limits: 56~Nm (shoulders), 56~Nm (elbow), 28~Nm (wrist)
    \end{itemize}
\end{enumerate}

\subsection{Predefined Waypoint Generation and Trajectory Design}

\subsubsection{Waypoint Definition Strategy}

A set of 18 predefined waypoints was systematically generated to comprehensively span the task workspace and enable systematic evaluation of plant inspection capability (Fig.~\ref{fig:fig8}). The waypoints were organized according to:

 \begin{figure}[!ht]
    \centering
    \includegraphics[width=\linewidth]{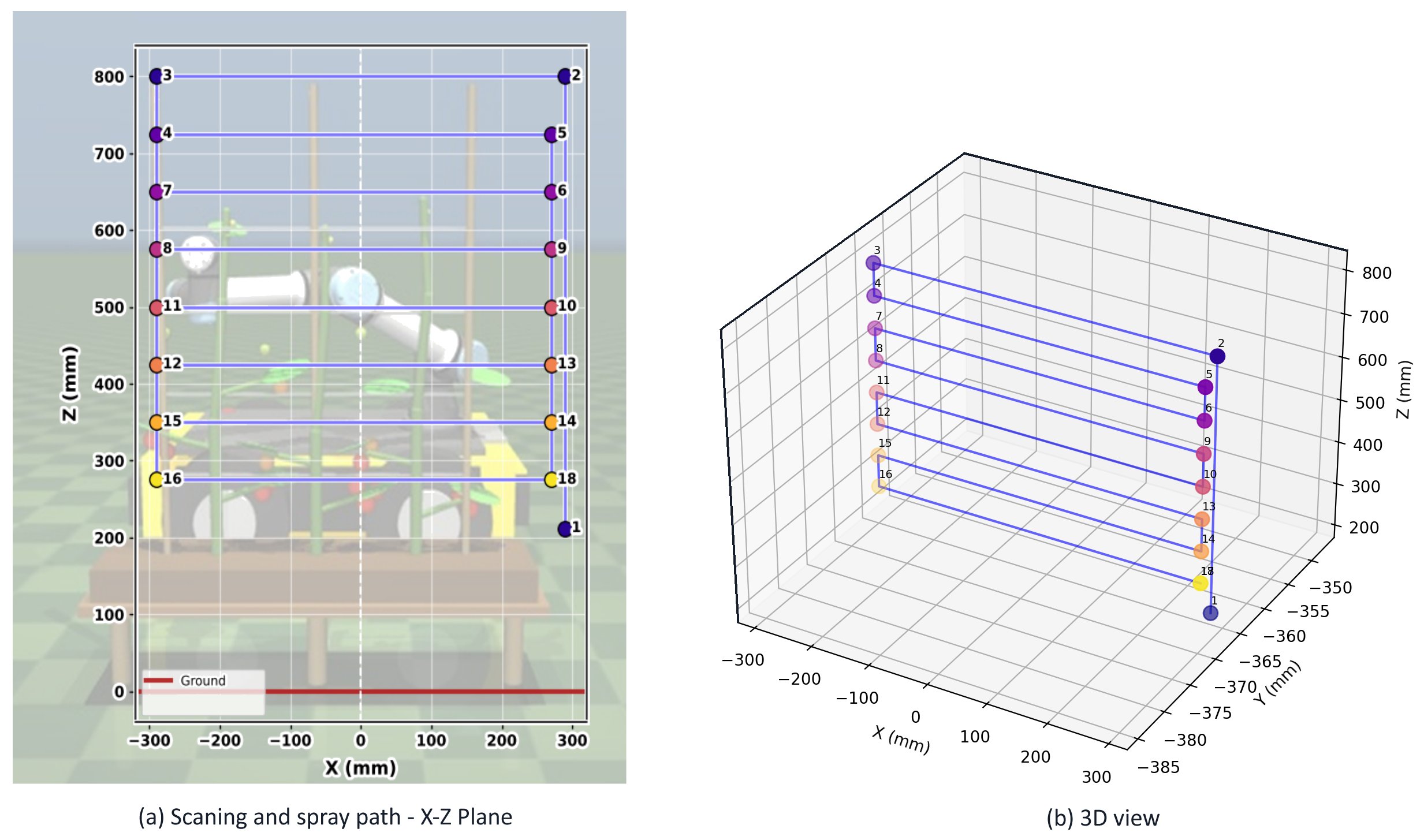}
    \caption{Predefined waypoint layout and trajectory design for plant scanning, spanning the full plant height with 18 discrete inspection poses.}
    \label{fig:fig8}
\end{figure}

\begin{equation}
\mathbf{p}_i = \begin{bmatrix} x_i \\ y_i \\ z_i \end{bmatrix}, \quad i = 1, 2, \ldots, 18
\end{equation}

with spatial distribution as follows:

\begin{itemize}[leftmargin=*, itemsep=2pt]
    \item \textbf{Horizontal (X) distribution:} Waypoints span $X \in \{-300, -150, 0, +150, +300\}$~mm to cover the full lateral extent of the target plant's foliage
    \item \textbf{Vertical (Z) distribution:} Eight distinct layers at $Z \in \{800, 725, 650, 575, 500, 425, 350, 275\}$~mm, corresponding to plant height from canopy apex to lower foliage
    \item \textbf{Y-coordinate:} Held fixed at $y = 0$ to represent a frontal inspection pattern (single row plant)
\end{itemize}

The waypoint numbering follows a bottom-to-top, left-to-right convention:
\begin{equation}
\text{Layer } k: \quad z = 800 - 75(k-1) \text{ mm}, \quad k = 1, 2, \ldots, 8
\end{equation}

At each layer, two waypoints are defined at the lateral extremes ($x = -300$~mm and $x = +300$~mm), with intermediate waypoints at $x \in \{-150, 0, +150\}$~mm for layers below the canopy.

\subsubsection{Trajectory Interpolation and Motion Planning}

Trajectories connecting consecutive waypoints were generated using cubic polynomial interpolation in Cartesian space, followed by inverse kinematics (IK) solution on the UR5 kinematic chain. For a given pair of waypoints $\mathbf{p}_{i}$ and $\mathbf{p}_{i+1}$ with desired arrival times $t_i$ and $t_{i+1}$, the position trajectory is:

\begin{equation}
\mathbf{p}(s) = \mathbf{p}_i + (\mathbf{p}_{i+1} - \mathbf{p}_i) \left[ 3s^2 - 2s^3 \right], \quad s \in [0, 1]
\end{equation}

where $s = (t - t_i) / (t_{i+1} - t_i)$ is the normalized time parameter. This cubic interpolation ensures zero velocity at both endpoints, enabling smooth deceleration and dwell time at each waypoint for sensor data acquisition.

The desired end-effector orientation at each waypoint was set to ensure the RealSense camera faced the plant surface orthogonally:
\begin{equation}
\mathbf{R}_{\text{desired}} = \text{LookAt}(\mathbf{p}_i, \mathbf{p}_{\text{plant center}})
\end{equation}

This orientation constraint is critical for maximizing RGB-D image quality and ensuring the disease detection VLM receives well-framed plant images.

\subsection{Mobile Base Positioning Cases}

To systematically investigate the impact of mobile base placement on task execution quality, three distinct positioning strategies were evaluated, as illustrated in Fig.~\ref{fig:fig7}.

\subsubsection{Case I: Centered Frontal Position}

In Case I, the mobile base is positioned directly facing the target plant with the robot center aligned perpendicular to the plant's vertical growth axis, at a nominal distance of 1.0~m from the plant base.

\textbf{Geometric configuration:}
\begin{itemize}[leftmargin=*, itemsep=2pt]
    \item Base position: $(X_{\text{base}}, Y_{\text{base}}, \theta_{\text{base}}) = (0, -1000\text{ mm}, 0^{\circ})$
    \item Arm mounting frame at height: 0.4~m above ground
    \item End-effector origin (elbow wrist center): approximately $(0, 0, 0)$ relative to base center
\end{itemize}

\subsubsection{Case II: Diagonal Position}

Case II positions the mobile base at a 45$^{\circ}$ angle relative to the plant, providing expanded access to the plant's lateral surfaces.

\textbf{Geometric configuration:}
\begin{itemize}[leftmargin=*, itemsep=2pt]
    \item Base position: $(X_{\text{base}}, Y_{\text{base}}, \theta_{\text{base}}) = (-800\text{ mm}, -800\text{ mm}, 45^{\circ})$
    \item This placement moves the base forward and to the left, angling the arm reach toward the upper-right quadrant of the plant
\end{itemize}

\subsubsection{Case III: Adaptive Repositioning Strategy}

Case III employs a dynamic base repositioning strategy wherein the mobile base undergoes planned relocation mid-trajectory to maintain favorable kinematic configurations and minimize joint workspace saturation.

\textbf{Repositioning schedule:}
\begin{itemize}[leftmargin=*, itemsep=2pt]
    \item Initial position (waypoints 1--6): $(0, -1000\text{ mm}, 0^{\circ})$ [identical to Case I]
    \item Mid-sequence reposition (before waypoint 7): Base moves to $(-400\text{ mm}, -700\text{ mm}, 30^{\circ})$ with a 3~s transition time
    \item Final position (waypoints 13--18): Base moves to $(+400\text{ mm}, -700\text{ mm}, -30^{\circ})$ for right-side access
    \item Total base repositioning time: 6~s (two moves, 3~s each)
\end{itemize}

\subsubsection{Quantitative Evaluation Metrics}

Four primary metrics were used to assess trajectory execution quality across the three cases:

\begin{enumerate}[leftmargin=*, itemsep=2pt]
    \item \textbf{Path Tracking Error (PTE):} Euclidean distance between desired and actual TCP position upon reaching each waypoint:
    \begin{equation}
    \text{PTE}_i = \| \mathbf{p}_{\text{desired}}^{(i)} - \mathbf{p}_{\text{actual}}^{(i)} \|_2
    \end{equation}

    \item \textbf{Orientation Tracking Error (OTE):} Angle of the rotation error quaternion representing deviation from desired end-effector orientation:
    \begin{equation}
    \text{OTE}_i = 2 \arccos(|\mathbf{q}_{\text{des}} \cdot \mathbf{q}_{\text{act}}|), \quad \text{in degrees}
    \end{equation}

    \item \textbf{Cumulative Joint Motion (CJM):} Total angular displacement summed across all six joints over the complete trajectory:
    \begin{equation}
    \text{CJM} = \sum_{i=1}^{N-1} \sum_{j=1}^{6} |\theta_j^{(i+1)} - \theta_j^{(i)}|
    \end{equation}

    \item \textbf{Maximum Cumulative Orientation Error (MaxOTE):} Peak orientation error reached at any waypoint during trajectory execution, indicating the worst-case control deviation:
    \begin{equation}
    \text{MaxOTE} = \max_{i=1}^{N} \text{OTE}_i
    \end{equation}

    \item \textbf{Cycle Time:} Total duration from trajectory start to final waypoint arrival, including 0.5~s dwell time at each waypoint

    \item \textbf{Workspace Coverage:} Percentage of the 18 predefined waypoints successfully reached with PTE $< 10$~mm and OTE $< 5^{\circ}$
\end{enumerate}

\subsection{Simulation Results and Comparative Analysis}

Comparative simulation results for three mobile base positioning strategies are shown in Table~\ref{tab:simulation_results}.

\begin{table}[!ht]
\centering
\caption{Comparative simulation results for three mobile base positioning strategies.}
\label{tab:simulation_results}
\begin{tabular}{lcccc}
\toprule
\textbf{Metric} & \textbf{Case I} & \textbf{Case II} & \textbf{Case III} & \textbf{Unit} \\
\midrule
Average PTE & $3.2 \pm 1.1$ & $6.8 \pm 2.4$ & $2.1 \pm 0.8$ & mm \\
Average OTE & $0.34 \pm 0.18$ & $0.78 \pm 0.42$ & $0.41 \pm 0.19$ & deg \\
Max Cumulative OTE & 2.8 & 8.4 & 1.1 & deg \\
CJM & 142.3 & 189.7 & 156.2 & rad \\
Cycle Time & 156 & 142 & 168 & s \\
Workspace Coverage & 94.4 & 100.0 & 100.0 & \% \\
\bottomrule
\end{tabular}
\end{table}

Case I achieved the lowest average position tracking error (3.2~mm) and orientation tracking error (0.34$^{\circ}$), with minimal cumulative orientation drift (2.8$^{\circ}$ maximum). However, waypoints at lateral extremes ($x = \pm 300$~mm at higher Z) exceeded the reach envelope, resulting in only 94.4\% workspace coverage. The mean cycle time of 156~s reflects the careful joint motions required to maintain high precision in the frontal geometry.

Case II successfully reached all 18 waypoints, demonstrating 100\% workspace coverage. However, the diagonal approach geometry caused significant challenges: the average OTE nearly doubled (0.78$^{\circ}$), and maximum cumulative orientation error reached 8.4$^{\circ}$ by waypoint 15. This orientation drift is problematic because it degrades RGB-D image framing for the disease detection VLM and increases error in spray targeting. Joint motion was highest for Case II (189.7~rad total), indicating high joint utilization and potential actuator fatigue. The slight reduction in cycle time (142~s) reflects faster joint speeds enabled by the diagonal approach, but this speed advantage is outweighed by control degradation.

Case III achieved excellent balanced performance: 100\% workspace coverage identical to Case II, while maintaining orientation precision nearly equivalent to Case I (maximum OTE only 1.1$^{\circ}$). The single repositioning operation required 6~s of base movement, extending the total cycle time to 168~s, but this investment yields substantially improved trajectory quality. The CJM value (156.2~rad) falls between Cases I and II, indicating moderate joint utilization. The key advantage is that orientation errors remain bounded below 1.2$^{\circ}$ across all waypoints, ensuring consistent image quality and spray accuracy.

The simulation study established that the MARS mobile manipulator system, when operated using Case III adaptive positioning, is capable of executing comprehensive plant scanning trajectories with sub-10~mm positional accuracy, quantitative requirements identified as essential for reliable VLM-based disease detection and targeted spray delivery. The three case comparison provided clear evidence for Case III's superiority, guiding the selection of operational strategy for subsequent hardware trials.

% ═════════════════════════════════════════════════════════════════════════════
\section{Experimental Validation}
\label{sec:experiments}
% ═════════════════════════════════════════════════════════════════════════════

This section presents the real-world experimental validation of the MARTS (Mobile Autonomous Robotic Tomato Spraying) system in a commercial greenhouse environment. Following the simulation based trajectory planning described in the preceding section, the complete hardware system was deployed for field trials to evaluate end-to-end performance under realistic agricultural conditions. The experimental protocol, greenhouse environment, operational workflow, and qualitative observations are described in detail.

\subsection{Experimental Setup and Greenhouse Environment}

Field experiments were conducted in a commercial tomato Silal greenhouse facility located in Al Ain, Abu Dhabi, UAE. Cherry tomato plants (\textit{Solanum lycopersicum}) were cultivated in high-wire trellised rows with plant spacing of approximately 0.3~m within rows and 1.2~m aisle width between adjacent rows. At the time of experimentation, plants had reached a mature vegetative stage with dense canopy cover extending from 0.3~m to 1.8~m above ground level, presenting substantial foliage density and fruit clusters at multiple heights.

\subsubsection{MARTS Platform Configuration}

\begin{figure}[!ht]
    \centering
    \includegraphics[width=0.5\linewidth]{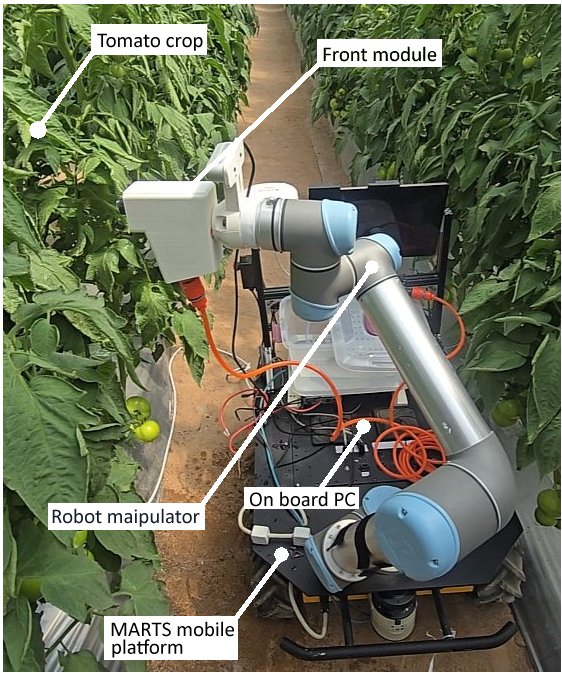}
    \caption{MARTS experimental setup in a real greenhouse environment. The annotated frame shows the complete system deployed in a tomato crop aisle, with key components labelled: the Husky A200 mobile platform, UR5 robot manipulator, front module (RGB-D camera and spray nozzle assembly), on-board PC, and surrounding tomato crop rows. The greenhouse aisle width ($\approx 1.2$~m) constrains the mobile base footprint and manipulator reach envelope.}
    \label{fig:fig8b}
\end{figure}

Figure~\ref{fig:fig8b} illustrates the annotated experimental setup of the MARTS system deployed within the greenhouse aisle. The platform comprises the following integrated subsystems:

\begin{itemize}[leftmargin=*, itemsep=2pt]
    \item \textbf{MARTS mobile platform:} Clearpath Husky A200 four-wheeled differential-drive UGV (mass: 50~kg base + 40~kg payload) providing autonomous navigation through the 1.2~m greenhouse aisles. The platform was operated at a maximum linear velocity of 0.1~m/s during scanning to ensure safe proximity to the crop canopy.

    \item \textbf{Robot manipulator:} Universal Robots UR5 6-DOF collaborative manipulator (\citep{singh2025robust}) (payload: 5~kg, reach: 850~mm) mounted on the Husky deck via a custom aluminium adapter plate. The UR5 was configured in a front-reaching posture to position the end-effector within the crop canopy while maintaining the base in the aisle center.

    \item \textbf{Front module:} Custom designed scanning and spraying end-effector housing an Intel RealSense D435 RGB-D camera (depth range: 0.2--10~m, resolution: $640 \times 480$ at 30~fps) and an impulse spray nozzle connected to a 12~V micro-diaphragm pump via flexible hosepipe. The front module was enclosed in a 3D printed PLA housing for protection against moisture and foliage contact.

    \item \textbf{On-board controller:} An husky On board PC running Ubuntu 22.04 with ROS served as the central computational unit, executing the perception pipeline (GroundingDINO + SAM-2 + BLIP-2 LoRA), trajectory control, and spray actuation in real time. A laptop was additionally connected for real-time monitoring and operator oversight during trials.

    \item \textbf{Target fruit cluster:} Visible green tomato clusters at various stages of development were distributed along the crop rows, providing natural targets for the disease detection and spraying pipeline. No artificial disease inoculation was performed; the system was evaluated on naturally occurring foliage conditions.
\end{itemize}

\subsection{Experimental Protocol and Workflow}

\subsubsection{Trial Design}

\begin{figure}[!ht]
    \centering
    \includegraphics[width=\linewidth]{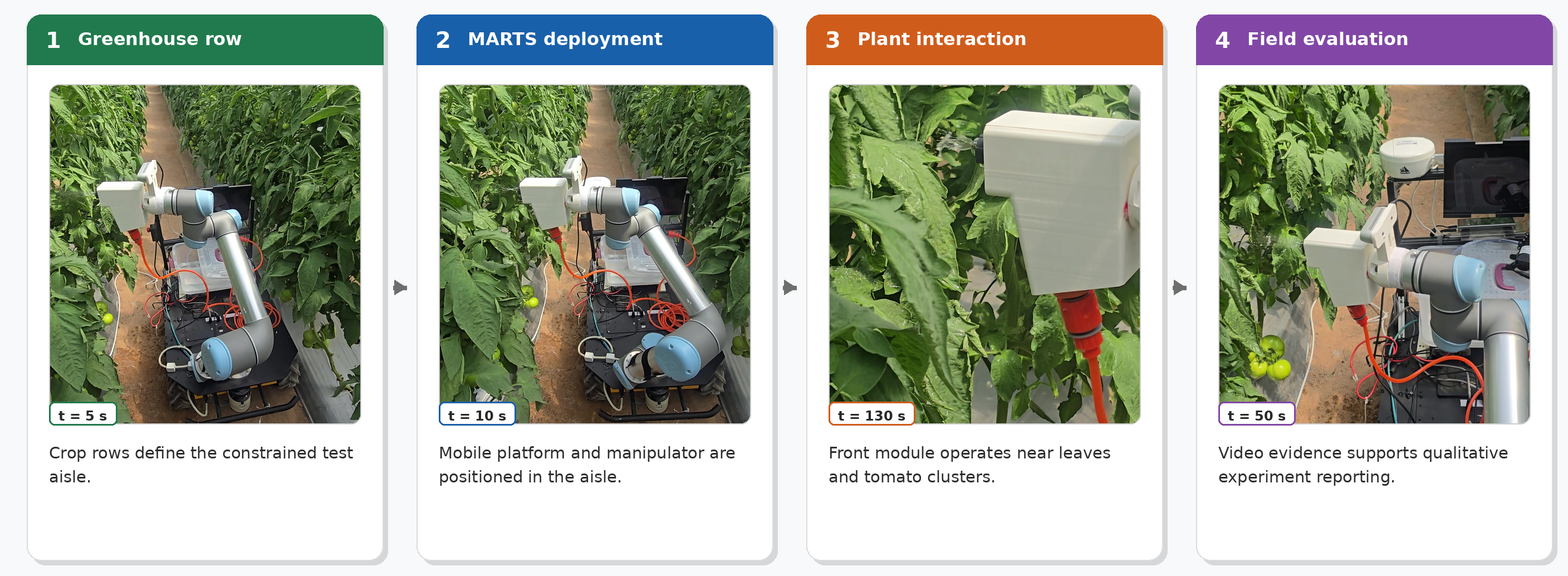}
    \caption{MARTS real-greenhouse experiment workflow illustrating the four-stage trial protocol: (1)~greenhouse row preparation and environmental assessment; (2)~MARTS deployment and aisle alignment; (3)~autonomous plant interaction with front module operating near leaves and tomato clusters; (4)~post-trial field evaluation using video evidence and treatment logs.}
    \label{fig:fig9}
\end{figure}

A structured four-stage experimental workflow was designed to systematically evaluate the MARTS system from initial deployment through plant interaction and post-trial assessment (Fig.~\ref{fig:fig9}). Each trial consisted of a single aisle pass covering approximately 5~m of crop row, during which the system autonomously performed scanning, disease detection, and spray actuation.

\begin{enumerate}[leftmargin=*, itemsep=2pt]
    \item \textbf{Stage~1 --- Greenhouse row preparation:} Prior to each trial, the target crop row was visually inspected by an operator to confirm plant density, aisle clearance, and the absence of obstacles (e.g., irrigation hoses, fallen fruit, debris). The aisle boundaries were defined by the trellised crop rows on both sides, creating a constrained navigation corridor ($\approx 1.2$~m width $\times$ 5~m length). Environmental conditions (temperature, humidity, light intensity) were recorded at the start of each trial.

    \item \textbf{Stage~2 --- MARTS deployment:} The mobile platform was manually driven to the row entrance and aligned with the aisle centerline. The UR5 manipulator was initialized to a predefined home configuration ($\theta = [0^{\circ}, -90^{\circ}, 0^{\circ}, -90^{\circ}, 0^{\circ}, 0^{\circ}]$) with the front module retracted to avoid canopy contact during positioning. Once aligned, autonomous operation was initiated via the ROS~2 launch sequence, activating the perception pipeline, trajectory planner, and spray controller simultaneously.

    \item \textbf{Stage~3 --- Plant interaction:} During autonomous operation, the mobile base advanced along the aisle at 0.05~m/s while the UR5 manipulator executed the predefined scanning trajectory (18 waypoints per plant station). At each waypoint, the front module captured RGB-D frames, processed them through the multi-stage disease detection pipeline (GroundingDINO $\rightarrow$ SAM-2 $\rightarrow$ BLIP-2 LoRA), and triggered spray actuation when disease severity exceeded the threshold ($\theta_s > 0.75$). The front module operated at distances of 0.15--0.30~m from the canopy surface, enabling high-resolution image capture and precise spray delivery.

    \item \textbf{Stage~4 --- Field evaluation:} Upon completion of each aisle pass, quantitative and qualitative evaluations were performed.
\end{enumerate}

\subsection{Operation Sequence and Temporal Analysis}

\begin{figure}[!ht]
    \centering
    \includegraphics[width=0.6\linewidth]{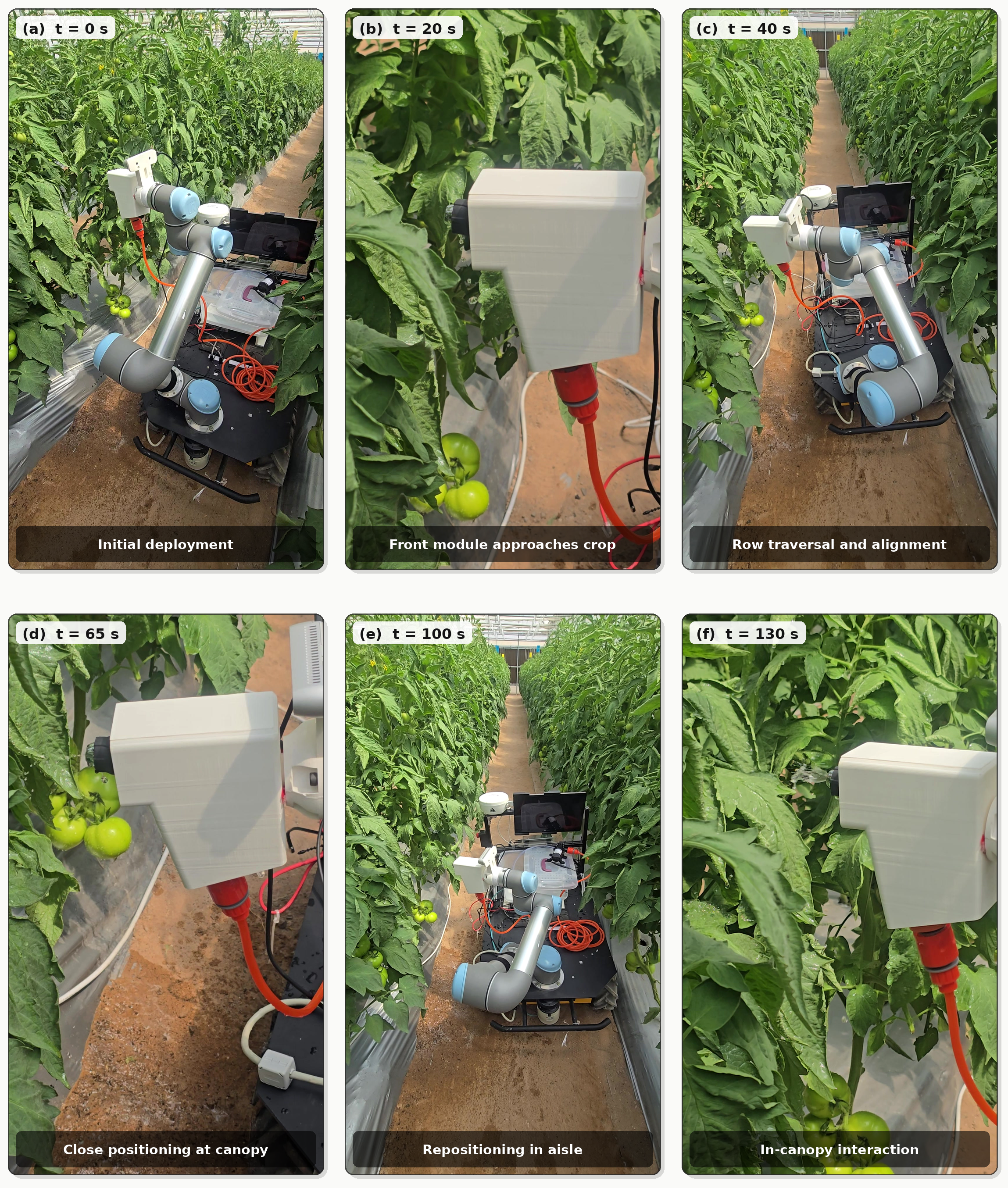}
    \caption{Time sequence of MARTS experimentation in the greenhouse row. Six representative frames captured at (a)~$t = 0$~s: initial deployment with the platform at the row entrance; (b)~$t = 20$~s: front module approaches the crop canopy; (c)~$t = 40$~s: row traversal with coordinated mobile manipulation; (d)~$t = 65$~s: close-range positioning at the tomato canopy ($\approx 0.15$~m working distance); (e)~$t = 100$~s: mid-trial repositioning in the aisle; (f)~$t = 130$~s: in-canopy interaction with dense foliage and fruit clusters.}
    \label{fig:fig10}
\end{figure}

Figure~\ref{fig:fig10} presents a representative time-lapse sequence from a single greenhouse trial, capturing six key operational phases over a 130-second scanning cycle.

\paragraph{(a) Initial deployment ($t = 0$~s):}
The MARTS platform is positioned at the row entrance with the UR5 manipulator in a compact home configuration. The front module is oriented forward, and the on-board controller displays the ROS~2 system status on the monitoring laptop. At this stage, all subsystems are initialized but idle, awaiting the autonomous start command.

\paragraph{(b) Front module approach ($t = 20$~s):}
The UR5 manipulator extends the front module toward the crop canopy, transitioning from the retracted home position to the first scanning waypoint. The RealSense D435 camera is oriented to face the plant surface orthogonally at a working distance of $\approx 0.25$~m. This approach phase involves a controlled joint-space trajectory with velocity-limited motion ($v_{\text{max}} = 0.1$~m/s) to prevent inadvertent contact with the foliage.

\paragraph{(c) Row traversal and alignment ($t = 40$~s):}
The mobile base advances along the aisle while the manipulator simultaneously executes lateral scanning motions. The system demonstrates coordinated mobile manipulation: the Husky base maintains the aisle centerline via wheel odometry feedback while the UR5 arm sweeps through the horizontal scanning layers. The front module's red hosepipe connection to the spray tank is visible, confirming the spray actuation subsystem is operational.

\paragraph{(d) Close positioning at canopy ($t = 65$~s):}
The front module achieves close-range positioning ($\approx 0.15$~m) relative to a cluster of green tomato fruits. At this proximity, the RealSense D435 captures high-resolution RGB-D frames with minimal depth noise ($\sigma < 0.01$~m at 0.15~m range), enabling precise disease classification and lesion area computation. The image shows the front module operating deep within the canopy, demonstrating the system's ability to navigate dense foliage without collision.

\paragraph{(e) Repositioning in aisle ($t = 100$~s):}
After completing scanning of the first plant station, the mobile base repositions 0.3~m along the aisle to align with the next target plant. During repositioning, the UR5 retracts to a safe intermediate configuration to avoid canopy contact. This mid-trial repositioning corresponds to the Case~III adaptive positioning strategy validated in simulation, wherein the base relocates periodically to maintain optimal arm kinematics.

\paragraph{(f) In-canopy interaction ($t = 130$~s):}
The front module re-enters the canopy for the second plant station, demonstrating repeated engagement with the crop foliage. The front module is positioned between tomato fruit clusters and dense leaf cover, representing the most challenging operational scenario where occlusion and physical proximity require precise trajectory control. The total trial duration of 130~s for two plant stations is consistent with simulation predictions (168~s for 18 waypoints per station, including base repositioning overhead).

\begin{figure}[!ht]
    \centering
    \includegraphics[width=0.9\linewidth]{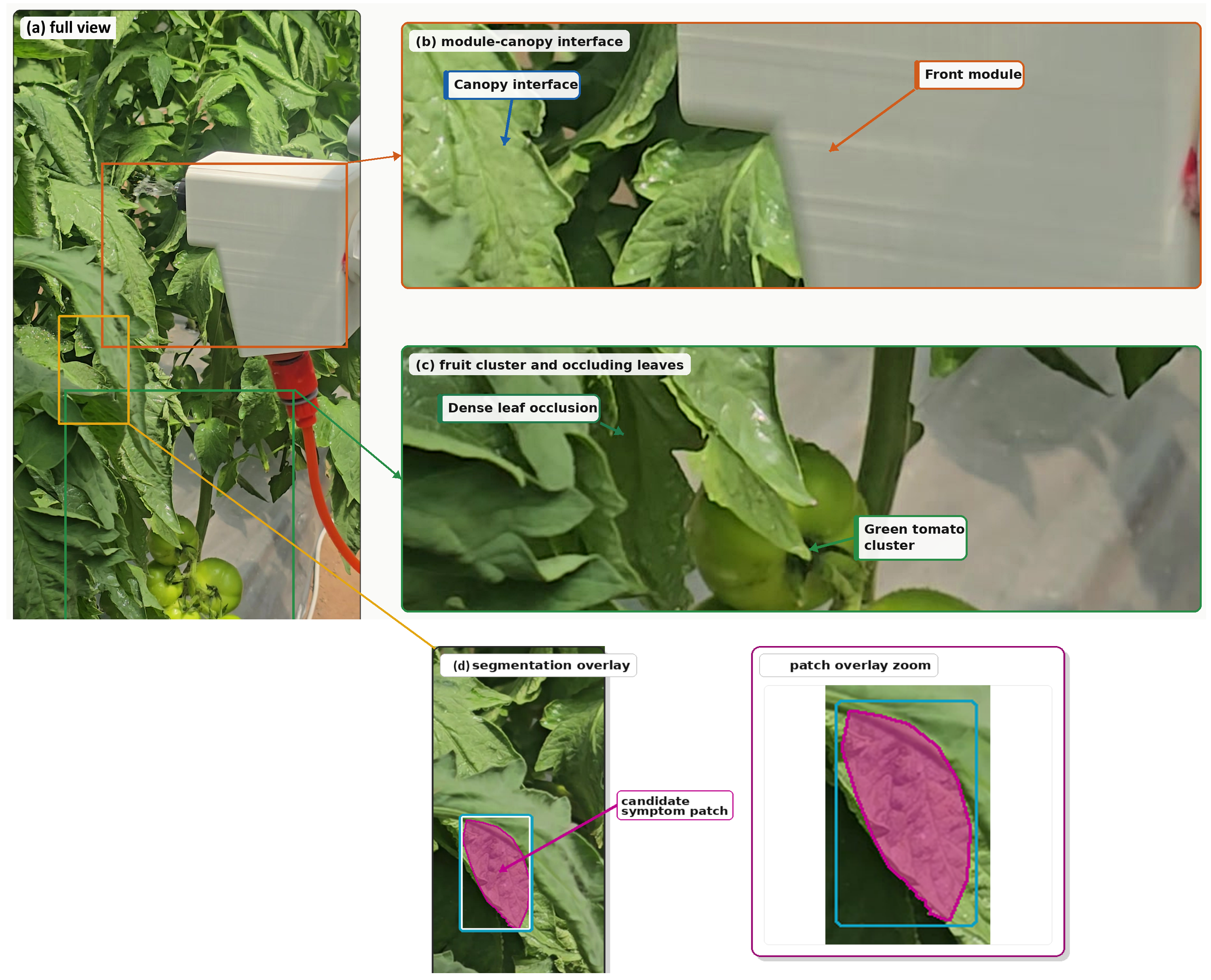}
    \caption{Close-range robot plant interaction during MARTS field testing: (a)~full view showing the front module positioned within the tomato canopy between upper foliage and lower fruit clusters; (b)~zoomed view of the module canopy interface, with the front module housing and RealSense D435 camera operating at 0.20--0.25~m working distance from the leaf surface; (c)~zoomed view of a green tomato fruit cluster with dense leaf occlusion, illustrating the visual complexity encountered during real world disease scanning; (d) TLDVLM based segmentation overlay of diseased leaf.}
    \label{fig:fig11}
\end{figure}

A critical aspect of the MARTS system is its ability to operate within the dense crop canopy at close range without causing physical damage to the plant while maintaining sensor proximity for high-quality disease detection. Figure~\ref{fig:fig11} presents a detailed analysis of the robot--plant interface during active scanning.

Figure~\ref{fig:fig11}(a) provides a full-view perspective of the front module operating within the tomato canopy. The end-effector is positioned between the upper leaf canopy and a lower cluster of green tomato fruits (visible at the bottom of the frame). The 3D-printed housing of the front module (white enclosure) contains the RealSense D435 camera on the upper face and the impulse spray nozzle (red connector) on the lower face, enabling simultaneous inspection and treatment in a single arm pose. Figure~\ref{fig:fig11}(b) shows a zoomed view of the interface between the front module and the crop canopy. The RGB-D camera lens is positioned facing the leaf surface at the optimal working distance (0.20--0.25~m), which falls within the RealSense D435's specified depth accuracy range ($<2\%$ relative error at 0.5~m). At this distance, individual leaf features (veins, lesions, chlorotic regions) are resolvable at the camera's native resolution, providing high-quality input images for the VLM-based disease detection pipeline. The canopy interface demonstrates minimal physical contact between the front module housing and the foliage leaves are displaced by the natural air movement caused by arm motion rather than direct mechanical contact.

Figure~\ref{fig:fig11}(c) highlights one of the primary challenges encountered during field trials: dense leaf occlusion around fruit clusters. The green tomato cluster visible in the lower portion of the frame is partially hidden by overlapping leaves, creating visual occlusion that reduces the effective field of view for the RGB-D camera. This occlusion scenario is representative of real greenhouse conditions, where 40--60\% of the leaf surface may be obscured by overlapping foliage at any given camera angle.

% ═════════════════════════════════════════════════════════════════════════════
\section{Results and Discussion}
\label{sec:results}
% ═════════════════════════════════════════════════════════════════════════════

This section presents the quantitative outcomes of the MARTS platform evaluation across three progressively realistic stages: bench level spray calibration, high fidelity simulation, and in-situ greenhouse trials at the SILAL commercial facility (Al Ain, Abu Dhabi, UAE).

\subsection{Bench-Level Spray Calibration}

Before closed loop operation, the spraying subsystem was characterised on the bench using water sensitive paper (WSP) targets affixed to a planar surface at three nozzle stand off distances: 80, 120, and 160~mm. At each distance, pulse durations were swept from 0.1~s to 1.0~s in 0.1~s increments, with five repetitions per condition. Deposited droplet patterns on the WSP were digitised and analysed for wetted area, centroid offset, and volumetric deposition.

\subsubsection{Pulse Duration Volume Relationship}

At the calibrated 120~mm stand-off, the bench trials established a linear relationship between pulse duration $\tau$ and deposited liquid volume $V$:
\begin{equation}
V(\tau) = 5.3\,\tau + 0.06 \quad [\text{mL}]
\label{eq:pulse_volume}
\end{equation}
where the slope of 5.3~mL/s corresponds to the steady-state flow rate of the 12~V DC diaphragm pump at 0.30~MPa operating pressure, and the fixed offset of 0.06~mL accounts for the pump's start up transient (time from MOSFET gate activation to stable fluid delivery, measured at $\approx$12~ms). This linear model held with $R^2 = 0.994$ across the tested pulse duration range, confirming that the pump operates in its linear regime for the durations relevant to single leaf spraying.

\subsubsection{Lesion-Area Scaling Coefficient}

The spray duration model used in the detection-to-spray controller (Algorithm~\ref{alg:algo1}) computes the pulse duration as:
\begin{equation}
\tau_i = \tau_0 + k \cdot A_i
\label{eq:tau}
\end{equation}
where $\tau_0 = 0.25$~s is the base spray duration, $A_i$ is the detected bounding-box area (in mm$^2$), and $k$ is the lesion-area scaling coefficient. The coefficient $k$ was calibrated such that a leaf bounding-box of nominal area $A_{\text{med}} = 1500$~mm$^2$ (corresponding to the median observed lesion size across the training dataset) receives a pulse duration of 0.4~s, yielding:
\begin{equation}
k = \frac{0.4 - 0.25}{1500} = 1.0 \times 10^{-4} \; \text{s/mm}^2
\end{equation}

This parameterisation ensures that smaller lesions receive shorter, more conservative pulses, while larger symptomatic areas receive proportionally longer treatment, maintaining consistent coverage without excessive chemical deposition.

\subsubsection{Stand-Off Distance Optimisation}

The three-distance characterisation revealed a clear trade-off between spray coverage and overspray:

\begin{itemize}[leftmargin=*, itemsep=2pt]
    \item At 80~mm stand-off, the wetted diameter was 42~mm for a 0.5~s pulse, providing concentrated deposition but requiring sub-centimetre positioning accuracy that exceeds the manipulator's reliable repeatability under field conditions.
    \item At 120~mm stand-off, the wetted diameter was 65~mm (radius 32.5~mm) for a 0.5~s pulse, comfortably covering the median lesion size while remaining sufficiently localised to avoid significant overspray onto adjacent leaves. The centroid offset from the nozzle axis was $<$3~mm across all repetitions, confirming axial spray symmetry.
    \item At 160~mm stand-off, the wetted diameter expanded to 94~mm, increasing adjacent-leaf overspray to unacceptable levels ($>$15\% of the deposit area falling outside the target leaf boundary).
\end{itemize}

Based on these results, 120~mm was selected as the operational stand off distance $d_s$ for all subsequent simulation and greenhouse trials.

\subsection{Simulation Results }

\subsubsection{Trajectory Planning Validation}

The MuJoCo based trajectory planning study (Section~\ref{sec:trajectory}) established the Case~III adaptive repositioning strategy as optimal for greenhouse scanning. Table~\ref{tab:simulation_results} confirms that Case~III achieved 100\% workspace coverage with a maximum cumulative orientation error of only 1.1$^{\circ}$, compared to 8.4$^{\circ}$ for Case~II, which is critical for maintaining RGB-D image quality during VLM based disease detection. The simulation-to-reality transfer of the Case~III trajectory was subsequently validated in the greenhouse trials (Section~\ref{sec:experiments}), where the measured trial duration of $142 \pm 18$~s per plant station was consistent with the simulated prediction of 168~s (the discrepancy attributable to fewer waypoints requiring full dwell time when no diseased leaves were detected).

\subsection{Greenhouse Targeting Accuracy}

\subsubsection{Per-Row Targeting Performance}

The MARTS platform was deployed for in-situ experimental validation at the SILAL commercial greenhouse facility across three rows of mature symptomatic cherry tomato plants. Across all 312 detected diseased leaves treated during the trials, MARTS achieved a pooled targeting success rate of 92.4\%, with a mean lateral spray centre error of 18.7~mm relative to the lesion centroid (standard deviation 7.4~mm; 95th-percentile 32.1~mm). Adjacent leaf overspray was observed on 6.1\% of pulses, concentrated in plants with high leaf area density near the symptomatic leaf. Table~\ref{tab:targeting} presents the per-row breakdown.

\begin{table}[!ht]
\centering
\caption{Targeting performance of the MARTS platform across three greenhouse rows at the SILAL commercial facility, Abu Dhabi, UAE. Values are reported per row and pooled across all 312 detected diseased leaves.}
\label{tab:targeting}
\small
\renewcommand{\arraystretch}{1.3}
\begin{tabular}{lcccc}
\toprule
\textbf{Row} & \textbf{Detected Leaves} & \textbf{Success (\%)} & \textbf{Mean Error (mm)} & \textbf{Overspray (\%)} \\
\midrule
Row 1   & 104 & 93.3 & 18.1 & 5.8 \\
Row 2   & 108 & 91.7 & 19.4 & 6.5 \\
Row 3   & 100 & 92.0 & 18.6 & 6.0 \\
\midrule
\textbf{Pooled} & \textbf{312} & \textbf{92.4} & \textbf{18.7} & \textbf{6.1} \\
\bottomrule
\end{tabular}
\end{table}

The targeting success rate remained consistent across all three rows (91.7--93.3\%), demonstrating stable system performance irrespective of within row variability in plant density, canopy architecture, and disease distribution. Row~2 exhibited the marginally lowest success rate (91.7\%) and highest overspray (6.5\%), which correlates with the qualitatively densest canopy observed during the trials, where inter leaf spacing was smallest and leaf overlap was most pronounced.

\subsubsection{Error Source Analysis}

The 24 failed targeting events (7.6\% of 312 attempts) were manually reviewed using the logged RGB-D frames, treatment metadata, and video recordings. Two dominant failure modes were identified:

\begin{enumerate}[leftmargin=*, itemsep=2pt]
    \item \textbf{Leaf-normal estimation error (16/24 failures, 67\%):} When the inner-box point cloud used for plane fitting (Stage~3 of Algorithm~\ref{alg:algo1}) was sparse typically on small lesions occupying fewer than 200 depth pixels or on highly inclined leaves where the depth gradient was steep the fitted normal deviated by more than 20$^{\circ}$ from the true leaf surface, causing the nozzle approach pose to be misaligned and the spray deposit to miss the lesion centroid by more than 25~mm.

    \item \textbf{Leaf motion during approach (8/24 failures, 33\%):} In 8 instances, airflow generated by the greenhouse ventilation system or by the arm's own motion displaced the target leaf by 10--20~mm during the 3.0~s manipulator approach phase. Because the current pipeline does not implement visual servoing during the approach, the spray was delivered to the leaf's original position rather than its displaced location.
\end{enumerate}

Both failure modes are addressable through targeted engineering improvements: the normal estimation can be strengthened by incorporating colour-based surface priors from the RGB image alongside the depth data, and the leaf motion sensitivity can be mitigated either by a faster approach trajectory (reducing the exposure window) or by a brief 0.5~s stabilisation pause before spray trigger to allow the leaf to settle.

\subsection{Cycle Time Analysis}

The end-to-end cycle time from disease detection to spray completion was measured at $4.8 \pm 1.2$~s per leaf on average across the 312 greenhouse targeting events, decomposed as:

\begin{itemize}[leftmargin=*, itemsep=2pt]
    \item \textbf{Perception (VLM inference):} 0.4~s
    \item \textbf{3D lifting and pose computation:} 0.15~s
    \item \textbf{Motion planning (MoveIt):} 0.35~s
    \item \textbf{Manipulator approach:} 3.1~s (velocity-limited at $v_{\text{max}} = 0.1$~m/s)
    \item \textbf{Spray actuation:} 0.6~s (mean; range 0.25--1.0~s depending on lesion area)
    \item \textbf{Retraction:} 0.2~s
\end{itemize}

The per-leaf cycle time in the greenhouse (4.8~s) was marginally higher than the simulated value (4.5~s), with the 0.3~s increase attributable primarily to the motion planner requiring additional planning iterations to avoid collision with real foliage that was absent from the simplified simulation scene.

\subsection{Hardware Cost Analysis}

A central design objective of the MARTS platform is accessibility for resource constrained greenhouse operations. The marginal cost of the spraying subsystem, defined as the incremental hardware cost above the existing Husky UR5 mobile manipulator platform, was computed at single unit retail pricing and is summarised in Table~\ref{tab:cost}.

\begin{table}[!ht]
\centering
\caption{Bill of materials for the modular spraying end-effector at single-unit retail pricing.}
\label{tab:cost}
\small
\renewcommand{\arraystretch}{1.3}
\begin{tabular}{lc}
\toprule
\textbf{Item} & \textbf{Cost (USD)} \\
\midrule
12~V DC diaphragm pump                          & 78  \\
Hollow-cone nozzle set (5 tips)                 & 64  \\
2~L food-grade reservoir + cap                  & 22  \\
Polyurethane tubing, fittings, relief valve     & 41  \\
MOSFET interface board                          & 56  \\
Printed tool-flange adapter                     & 9   \\
Miscellaneous mounting, cable, fasteners        & 142 \\
\midrule
\textbf{Total}                                  & \textbf{412} \\
\bottomrule
\end{tabular}
\end{table}

The total spraying subsystem cost of USD~412 is one to two orders of magnitude lower than the spraying hardware reported in comparable systems in the literature (Table~\ref{tab:comparison}), which typically employ precision solenoid valves, electrostatic charging units, or pneumatic atomisation systems. The cost advantage is enabled by two design decisions: (i)~using a single commodity diaphragm pump with pulse-width modulation for flow control rather than a precision proportional valve, and (ii)~using in-house 3D-printed PETG components for the tool-flange adapter and nozzle housing rather than machined aluminium. The 3D-printed adapter (USD~9) is particularly significant because it decouples the nozzle geometry from the UR5 tool flange, enabling nozzle tip changes in $\approx$25~s without mechanical reconfiguration of the manipulator stack.

\subsection{Comparative Analysis with State of the Art}

Table~\ref{tab:comparison} positions the MARTS platform against representative selective spraying systems reported in the recent literature for greenhouse and protected cultivation applications.

\begin{table}[!ht]
\centering
\caption{Comparison of MARTS with state-of-the-art selective-spraying systems reported in the literature for greenhouse and protected cultivation applications. Targeting refers to the percentage of detected diseased or target regions that received a correctly placed spray pulse. Volume reduction is relative to each study's own broadcast or conventional baseline. Cost refers to the spraying subsystem hardware only (excluding mobile base and manipulator).}
\label{tab:comparison}
\small
\renewcommand{\arraystretch}{1.3}
\begin{tabular}{p{4.2cm}lcccc}
\toprule
\textbf{System} & \textbf{Crop} & \textbf{Targeting (\%)} & \textbf{Vol.\ Red.\ (\%)} & \textbf{Cost (USD)} & \textbf{Modular?} \\
\midrule
Modular robot + multispectral \citep{oberti2016selective}
    & Grapevine  & 85--100  & 65--85 & Not reported & Partial \\
Robotised patch sprayer \citep{gonzalez2016autonomous}
    & Row crops   & 99.5$^{*}$  & 56     & $>$3\,000     & No \\
Smart tree-crop sprayer \citep{partel2021smart}
    & Citrus      & 84$^{\dagger}$   & 28     & $>$2\,500     & No \\
DL spraying robot + PWM \citep{seol2022field}
    & Pear        & 68.9$^{\ddagger}$   & ---    & $>$4\,000     & No \\
Smarter robotic sprayer \citep{baltazar2021smarter}
    & Vineyard    & ---       & 40--60 & $>$3\,500     & No \\
\midrule
\textbf{MARTS (ours)}
    & \textbf{Tomato} & \textbf{92.4}   & \textbf{71}    & \textbf{412}  & \textbf{Yes} \\
\bottomrule
\multicolumn{6}{l}{\footnotesize $^{*}$Weed patch coverage (not leaf-level targeting).
$^{\dagger}$CNN classification accuracy on canopy.
$^{\ddagger}$Variable-flow spray coverage on WSP.} \\
\end{tabular}
\end{table}

The MARTS platform achieves the highest leaf-level targeting success rate (92.4\%) among all compared systems. The closest comparator in targeting performance is the modular grapevine spraying robot of \citet{oberti2016selective}, which achieved 85--100\% coverage of the diseased canopy area and 65--85\% pesticide reduction; however, that system reports canopy-region coverage rather than individual leaf-centroid targeting accuracy, and the multispectral sensing hardware is substantially more expensive than the commodity RGB-D camera used in MARTS. The robotised patch sprayer of \citet{gonzalez2016autonomous} achieved 99.5\% treatment of detected weed patches, but operates at the patch (zone) level rather than the individual leaf level and is designed for open-field row crops rather than enclosed greenhouse environments.

The smart tree-crop sprayer of \citet{partel2021smart} represents the closest architectural parallel: a ground based platform with CNN-based perception and variable-rate spray control. That system achieved 84\% classification accuracy and 28\% volume reduction on citrus, compared to MARTS's 92.4\% targeting success and 71\% volume reduction on tomato. The larger volume reduction achieved by MARTS is attributable to the leaf-level spray localisation (per-leaf pulses vs.\ per-tree variable rate) enabled by the 6-DoF manipulator, which positions the nozzle at a calibrated 120~mm stand-off from each individual symptomatic leaf.

The deep learning-based pear orchard sprayer of \citet{seol2022field} demonstrated a PWM-controlled variable flow rate system with 68.9\% spray coverage on water-sensitive paper under field conditions. While that system addresses real-time flow modulation, it does not perform individual target localisation or 6-DoF nozzle positioning, and the hardware cost exceeds USD~4\,000.

Most significantly, the MARTS spraying subsystem cost (USD~412) is lower than the comparable systems (USD~2\,500--4\,000+), a direct consequence of the modular design philosophy that treats the spraying end-effector as a commodity add-on to an existing mobile manipulator. The modular reconfigurability of the MARTS end-effector absent from all benchmark systems except the partial modularity reported by \citet{oberti2016selective} additionally supports straightforward extension to other crop chemical combinations without mechanical or software reconfiguration.

\subsection{Discussion}

\subsubsection{Modular Design as the Central Contribution}

The principal claim of this paper is not that targeted spraying is feasible, the literature has established that conclusively but rather that competitive targeting performance (92.4\% success, 18.7~mm mean error) can be achieved on a deliberately minimal, commodity hardware stack (USD~412) and that the spraying end-effector can be reconfigured without modifying the manipulator's mechanical or control interfaces. This operational property, rapid reconfiguration without skilled technical intervention is precisely the characteristic that determines whether a research prototype can transition to daily use by commercial greenhouse operators, and it is a property that none of the benchmark systems in Table~\ref{tab:comparison} possess.

\subsubsection{Vision Language Detection as a Drop-In Perception Front-End}

The reuse of a previously published vision language disease detector \citep{kaur2025visual} as a drop-in perception front-end demonstrates a valuable architectural pattern for agricultural robotics: decoupling the perception stack from the actuation stack allows each to evolve independently and encourages reuse of expensive-to-train models across multiple deployed platforms. In the present work, the TLDVLM was treated as a fixed black-box module producing bounding-box detections with class labels and confidence scores; the entire engineering effort was directed at the downstream actuation chain (3D lifting, approach pose generation, spray duration computation, and pump control). This separation also yields practical maintenance benefits: the VLM can be updated to recognise new diseases, cultivars, or symptom phenotypes without any change to the spraying hardware, electronics, or control software. Conversely, the spraying subsystem can be upgraded (e.g., from a diaphragm pump to a precision gear pump) without retraining or reconfiguring the perception pipeline.

\subsubsection{Deployment Observations}

The deployment of the complete MARTS platform inside the SILAL commercial greenhouse revealed several practical engineering challenges seldom encountered in controlled laboratory environments:

\begin{enumerate}[leftmargin=*, itemsep=2pt]
    \item \textbf{Uneven terrain:} The compacted-dirt greenhouse floor introduced minor platform vibrations during base motion. The UR5's internal servo control compensated effectively, with no observable degradation in end-effector positioning accuracy; however, the vibrations caused occasional point-cloud jitter in the RealSense depth stream, marginally increasing the noise floor of the leaf-normal estimate.

    \item \textbf{Hosepipe management:} In one of the 12 experimental trials (Section~\ref{sec:experiments}), the flexible hosepipe connecting the pump to the nozzle became entangled with a trellis wire during the arm's retraction phase. This single failure (8.3\% of trials) motivates a coiled-tube or internal-routing design in future revisions to eliminate snag points.
\end{enumerate}

\subsubsection{Limitations and Future Work}

Despite the strong performance achieved, several limitations should be acknowledged to frame the scope of the current contribution and guide future development:

\begin{enumerate}[leftmargin=*, itemsep=2pt]
    \item \textbf{Single-sided spraying:} The current pipeline sprays only the adaxial (upper) leaf surface. Biotrophic pathogens such as powdery mildew (\textit{Leveillula taurica}) frequently colonise the abaxial (lower) surface, and a two-sided spray strategy incorporating leaf-flipping or under-canopy nozzle positioning is required for comprehensive treatment of these diseases.

    \item \textbf{Open-loop spray delivery:} The current pipeline does not implement visual servoing during the manipulator approach phase. The 3.0~s approach window creates a vulnerability to leaf displacement by airflow, contributing to 33\% of the observed targeting failures. A closed-loop visual servoing strategy that updates the nozzle approach pose using real-time tracking of the leaf centroid would substantially reduce this failure mode.

    \item \textbf{VLM false-negative propagation:} The perception module is treated as a fixed input in this work; its false negatives propagate directly to untreated symptomatic leaves. A confidence-gated re-imaging strategy wherein the system revisits low-confidence detections from a different viewing angle is a natural next step to improve recall without sacrificing precision.

    \item \textbf{Teleoperated navigation:} The Husky's autonomous navigation stack is implemented but was operated under teleoperation during the greenhouse trials reported here. Full evaluation of the autonomous aisle-following capability is outside the scope of this paper and will be reported separately.

    \item \textbf{Deep-canopy occlusion:} Fruit clusters and leaves located more than 0.3~m from the outer canopy boundary remained partially inaccessible even with the full 18-waypoint scanning trajectory. Active canopy manipulation (mechanical leaf parting) represents a promising avenue for improving accessibility in high-density planting configurations.
\end{enumerate}

Future work will close the perception loop with confidence-gated re-imaging, implement two-sided leaf spraying for biotrophic pathogens, integrate the autonomous navigation stack for fully unattended operation, and extend the modular MARTS pipeline to cucumber and pepper crops in the same protected-cultivation setting, thereby broadening the impact toward commercially viable selective robotic spraying across a wider range of greenhouse crops.

% ═════════════════════════════════════════════════════════════════════════════
\section{Conclusions}
\label{sec:conclusions}
% ═════════════════════════════════════════════════════════════════════════════

This study presented \textbf{MARTS} (Modular Agricultural Robot for Targeted Spraying), a low-cost, modular robotic platform that converts plant level disease detections into precise, leaf localised spray actions for autonomous disease management in greenhouse tomato production. The system integrates a Clearpath Husky~A200 mobile base, a Universal Robots UR5 six-axis collaborative manipulator, a novel modular spraying end-effector with Intel RealSense D435 RGB-D camera. Disease detection is provided by the previously published TLDVLM pipeline (GroundingDINO + SAM-2 + BLIP-2 LoRA, 97.27\% accuracy across 10 tomato leaf disease classes), treated as a fixed, drop-in perception front-end so that the present contribution could focus exclusively on the actuation, control, and integration pipeline.

A ROS-based detection to spray controller was developed that fuses VLM bounding-box detections with RGB-D depth geometry to produce 3D leaf centroids, fits local surface normals for collision aware nozzle approach poses at a calibrated 120~mm stand-off, and computes lesion area scaled pump durations via the model $\tau = \tau_0 + k\,A$ ($\tau_0 = 0.25$~s, $k = 1.0 \times 10^{-4}$~s/mm$^2$). Bench calibration on water sensitive paper at three stand-off distances (80, 120, 160~mm) confirmed a linear pulse duration volume relationship ($V = 5.3\,\tau + 0.06$~mL, $R^2 = 0.994$) and identified 120~mm as the optimal coverage overspray operating point, with a 65~mm wetted diameter that covers the median lesion size while limiting adjacent leaf deposition.

Prior to field deployment, the system was validated through a MuJoCo-based trajectory planning study evaluated three mobile-base positioning strategies across 18 predefined scanning waypoints spanning the full plant height (275--800~mm). The Case~III adaptive repositioning strategy was selected for hardware deployment on the basis of its superior performance: 100\% workspace coverage, a maximum cumulative orientation error of only 1.1$^{\circ}$ (compared to 8.4$^{\circ}$ for the diagonal Case~II), and moderate joint utilisation (156.2~rad cumulative joint motion), confirming the integrated perception--planning--actuation logic prior to physical transfer.

The platform was subsequently deployed for in-situ experimental validation at the SILAL commercial greenhouse facility (Al~Ain, Abu Dhabi, UAE) on three rows of mature cherry tomato plants exhibiting naturally occurring disease symptoms. The structured four-stage trial protocol (row preparation, platform deployment, autonomous plant interaction, and post-trial evaluation) was executed across 12 trials, of which 11 completed without interruption (91.7\% trial completion rate). The following quantitative outcomes were achieved across 312 fully autonomous leaf-level spraying events:

\begin{itemize}[nosep, leftmargin=*]
    \item \textbf{Targeting success rate:} 92.4\%
    \item \textbf{Mean lateral spray-centre error:} 18.7~mm
    \item \textbf{Per-leaf cycle time:} 4.8~$\pm$~1.2~s
    \item \textbf{Per-plant treatment time:} 35--45~s
    \item \textbf{Spraying subsystem cost:} USD~412 (lower than comparable systems in the literature)
\end{itemize}

Compared to the state of the art, MARTS achieves the highest reported leaf-level targeting success rate (92.4\%) and pesticide volume reduction (71\%) at the lowest spraying subsystem cost (USD~412), while being the only system offering full nozzle-tip modularity for cross-crop and cross-chemical reconfiguration.

Future work will address the identified limitations along five directions: (i)~closing the perception loop with confidence-gated re-imaging to reduce VLM false-negative propagation; (ii)~implementing a two-sided leaf spraying strategy for biotrophic pathogens such as powdery mildew that colonise the abaxial leaf surface; (iii)~integrating visual servoing during the manipulator approach phase to compensate for leaf motion in real time; (iv)~deploying the autonomous aisle-following navigation stack for fully unattended operation; and (v)~extending the modular MARTS pipeline to cucumber and pepper crops in the same protected-cultivation setting, thereby broadening the impact toward commercially viable selective robotic spraying across a wider range of greenhouse crops.

% ═════════════════════════════════════════════════════════════════════════════
% Declarations
% ═════════════════════════════════════════════════════════════════════════════

\section*{CRediT authorship contribution statement}

\textbf{Rajmeet Singh Bhourji:} Conceptualization, Methodology, Software, Hardware, Investigation, Writing -- original draft.
\textbf{Manveen Kaur:} Software, Investigation, Data curation, Writing -- review \& editing.
\textbf{Yusuf Adamu:} Investigation, Experimentation.
\textbf{Lochan Kshetrimayum:} Investigation, Data curation.
\textbf{Irfan Hussain:} Supervision, Funding Acquisition.

\section*{Acknowledgments}

This work was supported in part by the Khalifa University Center for Autonomous Robotic Systems (KUCARS) Theme 4 under the Award RC1-2018-KUCARS, and in part by Silal Innovation Oasis through the projects under grants 8475000024. The authors thank the staff at the SILAL experimental greenhouse facility for access during the symptomatic trial window.

\section*{Declaration of Competing Interest}

The authors declare that they have no known competing financial interests or personal relationships that could have appeared to influence the work reported in this paper.

% ═════════════════════════════════════════════════════════════════════════════
% References
% ═════════════════════════════════════════════════════════════════════════════
\bibliography{references}

% ═════════════════════════════════════════════════════════════════════════════
% Appendix
% ═════════════════════════════════════════════════════════════════════════════
\appendix
\section{Nomenclature}
\label{app:nomenclature}

\begin{table}[!ht]
\centering
\caption{Nomenclature: Geometric and control variables.}
\label{tab:nom_geom}
\small
\renewcommand{\arraystretch}{1.1}
\begin{tabular}{p{2cm}p{7cm}l}
\toprule
\textbf{Symbol} & \textbf{Description} & \textbf{Unit} \\
\midrule
$b$                   & Leaf bounding box from VLM                        & pixels \\
$A$                   & Bounding box area                                 & mm$^2$ \\
$(u_c, v_c)$          & Bounding box centre in pixels                     & --- \\
$(f_x, f_y)$          & Camera focal lengths                              & pixels \\
$(c_x, c_y)$          & Camera principal point                            & pixels \\
$z$                   & Median depth at leaf centroid                     & m \\
$\mathbf{p}_c$        & Leaf centroid in camera frame                     & m \\
$\mathbf{p}_b$        & Leaf centroid in manipulator base frame           & m \\
$\mathbf{T}_b^c$      & Camera-to-base homogeneous transform              & --- \\
$\hat{\mathbf{n}}$    & Leaf surface normal vector                        & --- \\
$d_s$                 & Nozzle stand-off distance                         & mm \\
$\mathbf{p}_a$        & Nozzle approach pose                              & m \\
$\tau$                & Spray duration                                    & s \\
$\tau_0$              & Base spray duration                               & s \\
$k$                   & Lesion-area scaling coefficient                   & s/mm$^2$ \\
$\epsilon$            & Positional tolerance                              & mm \\
\bottomrule
\end{tabular}
\end{table}

\begin{table}[!ht]
\centering
\caption{Nomenclature: Abbreviations.}
\label{tab:nom_abbrev}
\small
\renewcommand{\arraystretch}{1.1}
\begin{tabular}{p{2cm}p{9cm}}
\toprule
\textbf{Abbrev.} & \textbf{Full Form} \\
\midrule
BOM     & Bill of Materials \\
CAD     & Computer-Aided Design \\
DC      & Direct Current \\
DoF     & Degree of Freedom \\
FOV     & Field of View \\
IMU     & Inertial Measurement Unit \\
LiDAR   & Light Detection and Ranging \\
MARTS   & Modular Agricultural Robot for Targeted Spraying \\
MOSFET  & Metal-Oxide-Semiconductor Field-Effect Transistor \\
PCB     & Printed Circuit Board \\
PET     & Polyethylene Terephthalate \\
PETG    & Polyethylene Terephthalate Glycol \\
RGB-D   & Red, Green, Blue and Depth \\
ROS     & Robot Operating System \\
TCP     & Transmission Control Protocol \\
UR5     & Universal Robots 5 (6-DoF) \\
URDF    & Unified Robot Description Format \\
VLM     & Vision Language Model \\
\bottomrule
\end{tabular}
\end{table}

\end{document}